\documentclass[final]{cvpr}

\usepackage{times}
\usepackage{epsfig}
\usepackage{graphicx}
\usepackage{amsmath}
\usepackage{amssymb}
\usepackage[toc,title,page]{appendix}

\usepackage{multirow}
\usepackage{caption}
\usepackage{stfloats}
\usepackage{amsthm}
\usepackage{etoolbox}
\usepackage{xspace}
\usepackage{array,multirow}
\usepackage{boldline}
\usepackage[titlenumbered,ruled ]{algorithm2e}
\usepackage[symbol]{footmisc}
\usepackage{tablefootnote}
\usepackage[flushleft]{threeparttable}

\newcommand{\MyMapTemplatePrefixc}[4]{\expandafter#1\csname#3#4\endcsname{#2{#4}}} 
\forcsvlist{\MyMapTemplatePrefixc {\def} {\mathcal}{c}} {A,B,C,D,E,F,G,H,I,J,K,L,M,N,O,P,Q,R,S,T,U,V,W,X,Y,Z}  

\newcommand{\MyMapTemplatePrefixtb}[5]{\expandafter#1\csname#4#5\endcsname{#2{#3{#5}}}} 
\forcsvlist{\MyMapTemplatePrefixtb {\def} {\tilde}{\mathbf}{t}} {A,B,C,D,E,F,G,H,I,J,K,L,M,N,O,P,Q,R,S,T,U,V,W,X,Y,Z}  
\forcsvlist{\MyMapTemplatePrefixtb {\def} {\tilde}{\mathbf}{t}} {0,1,a,b,c,d,e,f,g,h,i,j,k,l,m,n,o,p,q,r,s,u,v,w,x,y,z}  
\makeatletter
\newcommand\footnoteref[1]{\protected@xdef\@thefnmark{\ref{#1}}\@footnotemark}
\makeatother
\newcommand{\MyMapTemplateNoPrefix}[3]{\expandafter#1\csname#3\endcsname{#2{#3}}}
\forcsvlist{\MyMapTemplateNoPrefix {\def} {\mathbf} } {0,1,a,b,c,d,e, f, g, h, i, j, k, l, m, n, o, p, q, r, u, v, w, x, y, z} 
\forcsvlist{\MyMapTemplateNoPrefix {\def} {\mathbf} } {A,B,C,D,E,F,G,H,I,J,K,L,M,N,O,P,Q,R,S,T,U,V,W,X,Y,Z}  

\usepackage[pagebackref=false,breaklinks=true,colorlinks,bookmarks=false]{hyperref}

\def\cvprPaperID{11326} 
\def\confYear{CVPR 2021}

\begin{document}

\title{Unbiased Mean Teacher for Cross-domain Object Detection}

\author{Jinhong Deng$^1$\space\space\space\space Wen Li$^{1}$\thanks{The corresponding author} \space\space\space\space Yuhua Chen$^2$\space\space\space\space Lixin Duan$^1$ \\
      $^1$University of Electronic Science and Technology of China $^2$Computer Vision Lab, ETH Zurich\\
      {\tt\small \{jhdeng1997, liwenbnu, lxduan\}@gmail.com, yuhua.chen@vision.ee.ethz.ch}
   }

\maketitle
\pagestyle{empty}
\thispagestyle{empty}

\begin{abstract}
   Cross-domain object detection is challenging, because object detection model is often vulnerable to data variance, especially to the considerable domain shift between two distinctive domains. In this paper, we propose a new Unbiased Mean Teacher (UMT) model for cross-domain object detection. We reveal that there often exists a considerable model bias for the simple mean teacher (MT) model in cross-domain scenarios, and eliminate the model bias with several simple yet highly effective strategies. In particular, for the teacher model, we propose a cross-domain distillation method for MT to maximally exploit the expertise of the teacher model. Moreover, for the student model, we alleviate its bias by augmenting training samples with pixel-level adaptation. Finally, for the teaching process, we employ an out-of-distribution estimation strategy to select samples that most fit the current model to further enhance the cross-domain distillation process. By tackling the model bias issue with these strategies, our UMT model achieves mAPs of $44.1\%$, $58.1\%$, $41.7\%$, and $43.1\%$ on benchmark datasets Clipart1k, Watercolor2k, Foggy Cityscapes, and Cityscapes, respectively, which outperforms the existing state-of-the-art results in notable margins. Our implementation is available at \href{https://github.com/kinredon/umt}{https://github.com/kinredon/umt}.
\end{abstract}

    \vspace{-2mm}
	\section{Introduction}
	
    In recent years, deep domain adaptation has gained increasing attention in computer vision community, due to that the supreme performances of deep models are normally restricted only in the domain of training data. When those trained models are applied to new environments, significant performance drops have often been observed in many computer vision tasks~\cite{chen2019learning,chen2018domain,ganin2016domain,saito2019strong,zhang2018collaborative}.

  In this work, we are specifically interested in the cross-domain object detection problem, because the strong demands from real-world scenarios. For instance, in autonomous driving, robust object detection is needed in different weathers and lighting conditions. Collecting annotations for all conditions can be extremely costly, and therefore models that can adapt to new environments without labeled data are highly desirable.
   
  Designing such domain adaptive detection models can be challenging. Compared to the image classification task, the output of object detection is richer and more complex, consisting of both the class labels and the bounding box locations. The two outputs are intrinsically coupled, making it more vulnerable towards data variance like scene changes, weather conditions, camera diversity, \textit{etc}. Various approaches have been proposed to address these issues, including instance and image-level adversarial training~\cite{chen2018domain}, strong and weak adversarial training~\cite{saito2019strong}, graph-based consistency~\cite{cai2019exploring}, \textit{etc}. 

  In this paper, we propose a new approach called Unbiased Mean Teacher (UMT) for cross-domain object detection. We build our approach based on the Mean Teacher (MT) model~\cite{tarvainen2017mean}, which is originally proposed for semi-supervised learning. By enforcing the consistency over perturbed unlabeled samples between the teacher and student models via distillation, it naturally gains improved robustness against data variance to some extent, and thus being used as our starting point for cross-domain object detection.

  However, solely using MT for cross-domain object detection often fails to produce promising results (see Section~\ref{sec:experiment} for detailed experimental study). We conduct a further experimental investigation into this issue and observe that there exists an essential model bias issue for MT in cross-domain scenarios. Specifically, in the presence of a large domain gap, the MT model can be easily biased towards the source domain, as the supervision is mainly from the source domain.
   
  To overcome such model bias occurred in the mean teacher model for cross-domain object detection, we design the unbiased mean teacher model with three strategies. Firstly, observing the biased teacher model often produces more precise predictions for the source images, we design a cross-domain distillation approach by using the source-like target images translated with CycleGAN~\cite{zhu2017unpaired} as the input for teacher model, and original target images as the input for student model. This significantly improves the effectiveness of distillation and reduces the influence of the model bias. Then, to further remedy the model bias of student model, we use the target-like source images as additional labeled data for training. Finally, for the teaching process, we also employ an out-of-distribution estimation strategy to identify the target samples that most fit the current MT model to enhance the cross-domain distillation process.

  The contributions of this work mainly can be summarized in three aspects: 1) \textbf{A new observation:} We reveal the essential model bias issue in the MT model for cross-domain object detection. 2) \textbf{A new model:} We propose a new domain adaptation framework called Unbiased Mean Teacher (UMT) for object detection, which addresses the model bias with several simple yet effective strategies. 3) \textbf{A new benchmark:} Our new UMT model achieves state-of-the-art performances on multiple datasets, setting up a new benchmark for cross-domain object detection research. Our UMT model achieves mAPs of $44.1\%$, $58.1\%$, $41.7\%$, and $43.1\%$ on benchmark datasets Clipart1k, Watercolor2k, Foggy Cityscapes, and Cityscapes, respectively.
	
	\section{Related Works}
	
	\noindent\textbf{Unsupervised Domain Adaptation:} 
	Unsupervised domain adaptation methods are designed to adapt a model from the labeled source domain to an unlabeled target domain. Many previous works aim to minimize the distance metric such as maximum mean discrepancy(MMD)~\cite{cariucci2017autodial,long2015learning,long2016unsupervised,long2017deep,sun2016deep}. Alternatively, adversarial training with domain classifier is also commonly used to learn domain-invariant representation~\cite{ganin2014unsupervised,ganin2016domain,sankaranarayanan2018generate,tzeng2017adversarial}. More similar to our work, French~\textit{et al.} proposed a model based on the Mean Teacher model~\cite{tarvainen2017mean}, and achieved state-of-the-art results on various benchmarks. Their model includes student and teacher models. The student model is trained using gradient descent while the weights of the teacher network are an exponential moving average of those of the student, and the inconsistency in predictions between the two models is penalized to encourage model robustness. The aforementioned domain adaptation models have focused on the task image classification, while in this work we study a more challenging object detection problem.

	\noindent\textbf{Object Detection:}
  Powered by the strong representation ability of deep convolution neuron network(CNN) models~\cite{he2016deep,krizhevsky2012imagenet,simonyan2014very}, object detection has made significant progress in recent years. Many DCNN-based methods have been proposed~\cite{cai2018cascade,girshick2015fast,girshick2014rich,he2015spatial,ren2015faster,redmon2016you,liu2016ssd,redmon2017yolo9000,shen2017dsod, lin2017feature, uijlings2013selective}, achieving remarkable performance in benchmarks such as PASCAL VOC~\cite{everingham2010pascal} and MSCOCO~\cite{lin2014microsoft}. Among all works, one of the most representative works is Faster RCNN~\cite{ren2015faster}, which extracts regions of interest~(RoIs) by a Region Proposal Network~(RPN), and then the prediction is made based on the feature sampled from the RoI. In this work, we test our model using Faster RCNN~\cite{ren2015faster} as our base detection network. But other detection networks should also be possible.
	
	\noindent\textbf{Cross Domain Object Detection:}
	Recently, many works have been proposed to address the domain shift problem occurred in object detection, using different techniques. Many approaches utilize adversarial learning manner with a gradient reverse layer to obtain domain-invariant feature, such as DA-Faster~\cite{chen2018domain}, SCDA~\cite{zhu2019adapting}, SWDA~\cite{saito2019strong}, SPLAT~\cite{tzeng2018splat}, MAF~\cite{he2019multi}, MDAL~\cite{xie2019multi}, CRDA~\cite{xu2020exploring}, CDN~\cite{su2020adapting}, HTCN~\cite{chen2020harmonizing}, \textit{etc}. MTOR~\cite{cai2019exploring} performs Mean Teacher~\cite{tarvainen2017mean} to explore object relation in region-level consistency, inter-graph consistency and intra-graph consistency. Shan~\textit{et al.}~\cite{shan2019pixel} employs generative adversarial network and cycle consistency for image translation in the pixel space and minimizes domain discrepancy in features. DM~\cite{kim2019diversify} yields various distinctive shifted domains from source domain and employs multi-domain-invariant representation learning to encourage features to be indistinguishable among the domains. The label-level adaptation~\cite{khodabandeh2019robust,kim2019self,roychowdhury2019automatic} has also been used for the task and produces improved detection performance. Our proposed UMT achieves improvements in large margins over the above methods. 
	
	\section{The Unbiased Mean Teacher Model}
	In this section, we start from the Mean Teacher~(MT) model~\cite{tarvainen2017mean} and discuss how to re-purpose it for cross-domain object detection. A simple MT model is firstly introduced by applying MT to object detection with necessary technical adjustment. Then, we give an analysis on the model bias problem of MT model in the cross-domain object detection task. Based on our analysis, we firstly address the model bias in teacher model through a cross-domain distillation for MT to maximally exploit the expertise of the teacher model and then propose to utilize the augmented training samples with pixel-level adaptation to further alleviate the bias in the student model. For the teaching process, we also employ an out-of-distribution detection strategy to select samples that most fit the current model to enhance the cross-domain distillation process. 
	
	\subsection{The Mean Teacher Model}
    Mean Teacher~(MT)~\cite{tarvainen2017mean} was initially proposed for semi-supervised learning. It consists of two models with identical architecture, a student model and a teacher model. The student model is trained using the labeled data as standard, and the teacher model uses the exponential moving average~(EMA) weights of the student model. Each sample prediction of the teacher model can be seen as an ensemble of the student model's current and earlier versions, therefore it is more robust and stable. By enforcing the consistency of teacher and student models using a distillation loss based on unlabeled samples, the student model is then guided to be more robust. The MT model has also been extended to unsupervised domain adaptation by using the target domain samples as the unlabeled data for distillation in~\cite{french2017self}.
    
    During the distillation, a small perturbation is added to the unlabeled data. By selecting samples with high probabilities for prediction, the student model is encouraged to reduce more variance on the unlabeled target samples, thus further enhancing the model robustness. Therefore it is suitable to be applied to object detection, for addressing the issue that object detection model is sensitive to data variance due to simultaneously predicting the tangled bounding boxes and object classes. 
    
    However, since MT does not explicitly address the domain shift, when applying the MT model to the cross-domain object detection task, the considerable domain shift might cause the predictions from teacher model unreliable, making the distillation less effective (see our investigation in Section~\ref{sec:model_bias}). The recent work~\cite{cai2019exploring} proposes to use the region graph to facilitate the distillation, however, it still does not directly address the intrinsic model bias of MT.

    \subsection{A Simple MT Model for Object Detection}
    \label{sec:simple_mt}
    \begin{figure}[t]

    \centering
    \includegraphics[width=.8\linewidth]{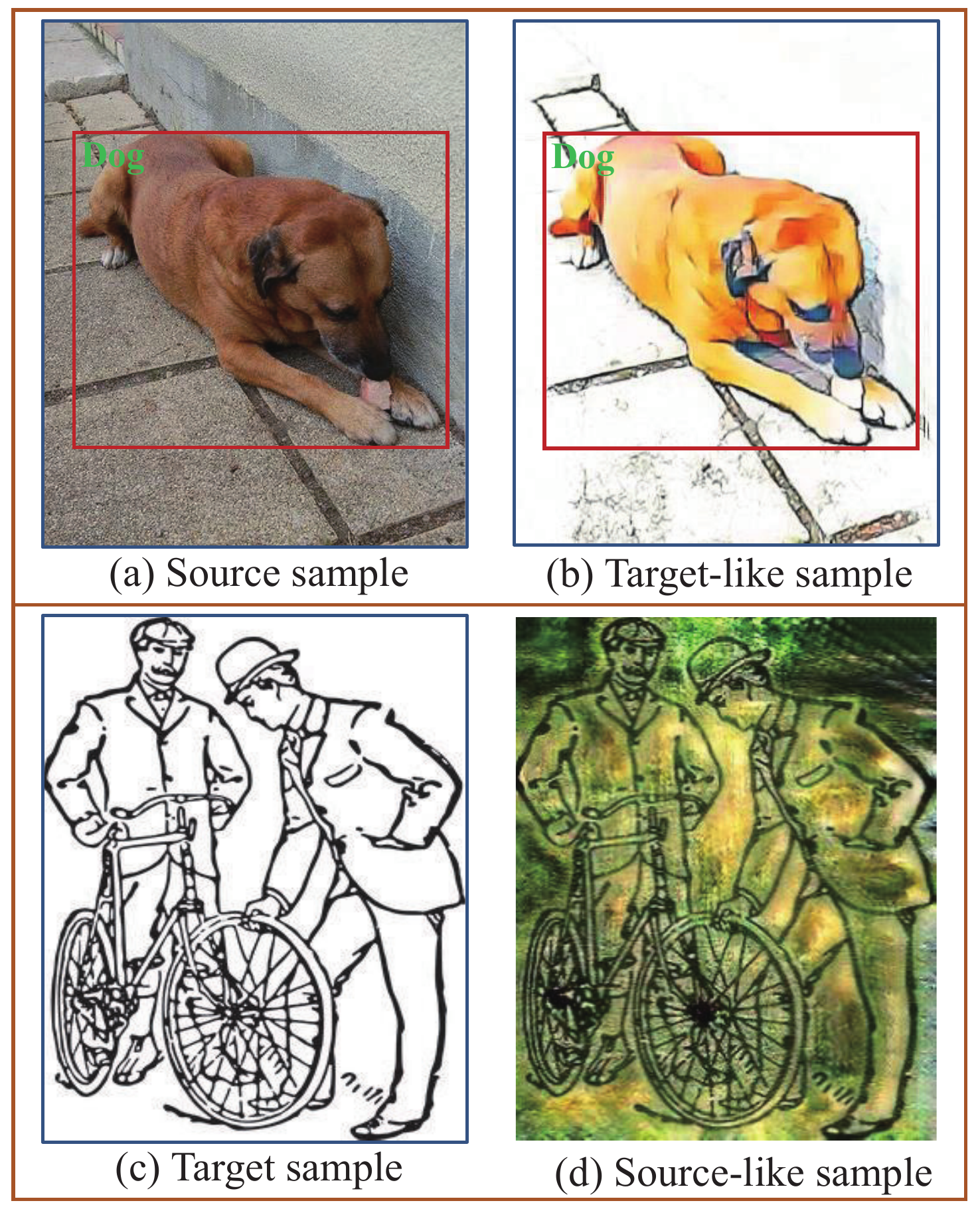}

   \caption{Examples of translated images. (a) and (c) are respectively a source example image from the PASCAL VOC dataset~\cite{everingham2010pascal}, and a target example image from the Clipart1k dataset~\cite{inoue2018cross}. (b) is the target-like image by translating the source image (a) into the target style, and (d) is the source-like image by translating the target image (c) into the source style.} 
   \label{fig:sample}
   \vspace{-4mm}
\end{figure}

    In the cross-domain object detection task, we have a set of source images annotated with object bounding boxes and their class labels, and a set of unlabeled target images. The source and target domain samples are drawn from different data distribution while their label space is the same. Its goal is to learn a model that could achieve good performance for the target domain.
    
    Formally, let us denote a source image as $\I^s$, which is annotated with multiple bounding boxes as well as their class labels. We denote by $\cB = \{B_j|_{j=1}^M\}$ as the set of bounding box coordinates with each $B_j = (x,y,w,h)$ representing a bounding box. Accordingly, we denote by $\cC = \{{c_j|_{j=1}^M}\}$ as the corresponding class labels, in which each  $c_j \in \{0, 1, \ldots, C\}$ corresponds to $B_j$ where $0$ stands for background, the others for object classes, and $C$ is the total number of classes. Then, the source domain can be represented as $\cD_s = \{(\I_{i}^{s}, \cB_{i}^{s}, \cC_{i}^s)|_{i=1}^{N_{s}}\}$, where $N_s$ is the number of source images. Similarly, the target domain can be defined as $\cD_t = \{\I_{i}^{t}|_{i=1}^{N_{t}}\}$ where $N_t$ is the number of target images. Note that labels for the target domain are not available.
    
    Following the protocol of mean teacher, firstly, the labeled source samples are passed through the student model for training. In particular, we employ Faster RCNN~\cite{ren2015faster} as our object detection backbone. Therefore the loss for training the student model with the labeled source samples can be written as:
   \begin{equation}
   \cL_{\text{det}}(\cB^s, \cC^s, \I^s) = \cL_{\text{rpn}}(\cB^s;\I^s) + \cL_{\text{roi}}(\cB^s, \cC^s;\I^s),
   \label{eqn:loss_det}
   \end{equation}
   where $\cL_{\text{rpn}}$ is the loss for the Region Proposal Network (RPN) module which is used for the candidate proposals generation, and $\cL_{\text{roi}}$ is the loss for the prediction branch which performs bounding box regression and classification. More details can be found in~\cite{ren2015faster}.

   Meanwhile, the unlabeled target samples are augmented with random cropping, padding and color jittering(i.e. brightness, contrast, hue and saturation augmentations). The augmented target samples are fed into teacher and student models, respectively, and then we take the instance predictions with high probabilities of teacher model to guide the student model via distillation. We denote the augmented target samples as $\hat{\I}^t$ for the student model, $\tilde{\I}^t$ for the teacher model. The loss for distillation can be defined as:
    \begin{equation}
    \begin{aligned}
   \mathcal{L}_{dist}(\tilde{\I}^t, \hat{\I}^t) = \cL_{\text{det}}(\cT_B(\tilde{\I}^t), \cT_C(\tilde{\I}^t), \hat{\I}^t),
    \end{aligned}
    \label{eqn:loss_dist}
   \end{equation}
    where $\mathcal{T}_B(\tilde{\I}_i^t)$ and $\mathcal{T}_C(\tilde{\I}_i^t)$ are the predicted bounding box coordinates and object classes with high maximum category score from the teacher model on the augmented image $\tilde{\I}_i^t$, and $\cL_{\text{det}}$ is the Faster RCNN loss defined in Eq.~(\ref{eqn:loss_det}).
    
    \begin{figure}[t]
   \centering
    \includegraphics[trim={0cm 8.5cm 16cm 0cm},clip,width=0.95\linewidth]{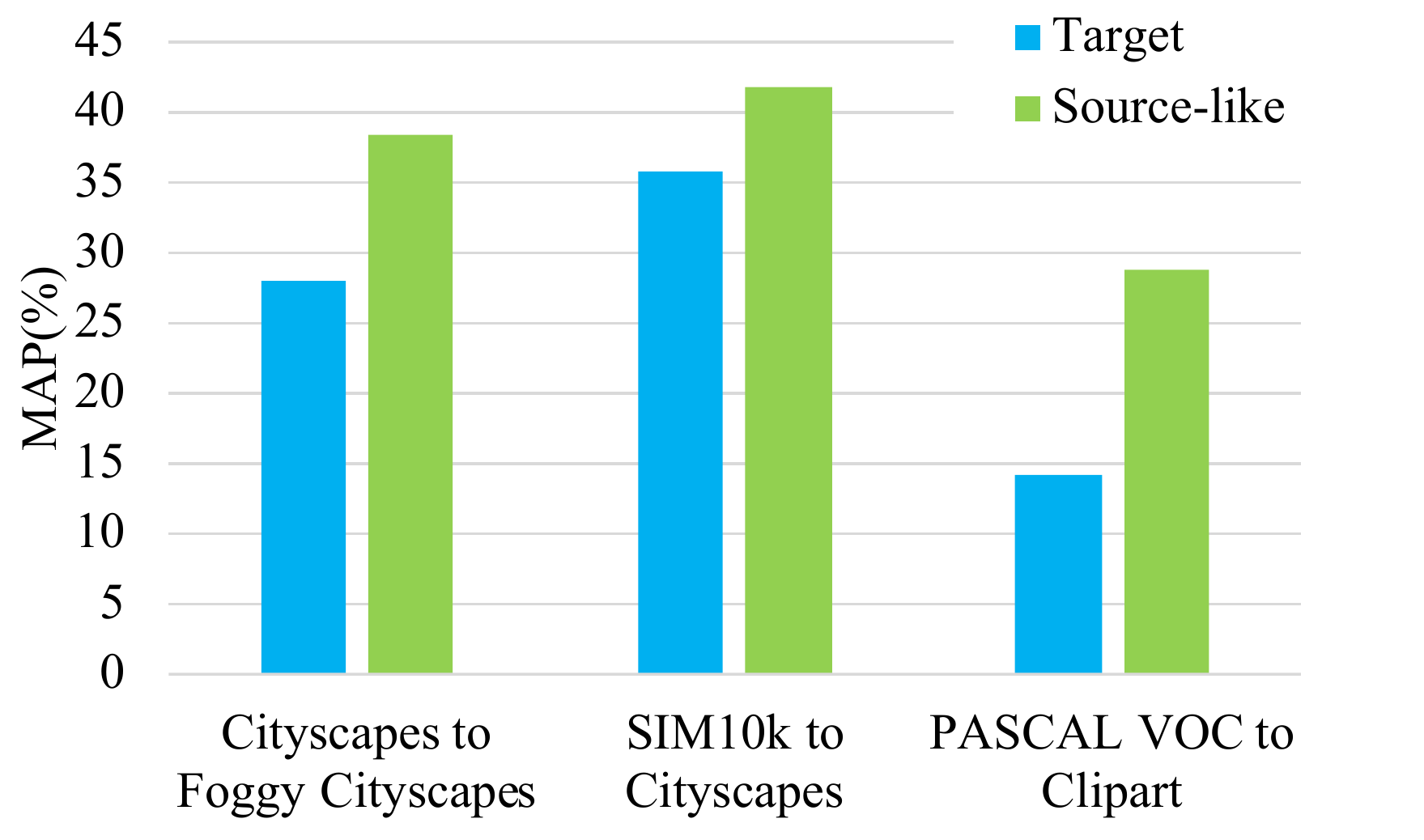}
   \caption{Evaluation of teacher models in mean teacher on target images and source-like images in different cross-domain object detection scenarios. Mean Average Precision (mAP) is used as the comparison metric. It can be observed that the teacher model produces significantly better detection results on source-like images, which clearly indicates the models are biased towards the source domain.}
   \label{fig:test_result}
   \vspace{-2mm}
   \end{figure}

    For obtaining $\mathcal{T}_B(\tilde{\I}_i^t)$ and $\mathcal{T}_C(\tilde{\I}_i^t)$, we firstly input $\tilde{\I}_i^t$ into the teacher model to get a group of bounding boxes and their class scores. Then, for each foreground category, we sort all proposals by their scores on this category and take Non-Maximum Suppression (NMS) to eliminate redundant proposals. Finally, we select the bounding boxes with their category score larger than a threshold $T$.
    Specifically, we denote $B_j$, $c_j$ and $p_j$ are the bounding boxes coordinates, category and maximum probability of the j-th instance predictions from the teacher model for an input image. The selected instance predictions can be defined as follows:
    \begin{equation}
    \begin{aligned}
   (\cT_B, \cT_C)=\left\{ (B_{j}, c_j) | p_{j} > T, \forall j \right \}.
    \end{aligned}
    \label{eqn:threshold}
    \vspace{-1mm}
   \end{equation}
  
	Finally, the overall loss of the mean teacher can be obtained by putting together those two losses:
   \begin{equation}
   \mathcal{L} = \cL_{\text{det}}(\cB^s, \cC^s, \I^s) + \lambda\mathcal{L}_{dist}(\tilde{\I}^t, \hat{\I}^t),
   \label{eqn:loss_mt}
   \end{equation}
   where $\lambda$ is a trade-off parameter.
\subsection{Investigating the Model Bias in Mean Teacher}
    \label{sec:model_bias}
   Although mean teacher can improve the robustness of predictions on the target domain, it is inevitable that learned models will be biased towards the source domain, as the supervision substantially comes from labeled source samples. As a result, in the distillation process, the prediction on the unlabeled target images produced by the teacher model could be deficient. It will degrade the object detection performance by a notable margin, especially for the unlabeled target domain. This issue might be relatively minor in the classification task. However, in object detection, minor biases in localization may cause a considerable difference in feature pooling and class prediction, and thus leads to inferior guidance to the student model.

	To verify this, we experiment on the mean teacher models trained in the aforementioned way on several cross-domain object detection datasets. In particular, we aim to investigate whether the teacher model is biased to the source domain, by comparing its performance on source and target samples. To ensure a fair comparison, we use the unpaired image-to-image translation by CycleGAN~\cite{zhu2017unpaired} to produce a source-like image for each target image. An example of the source-like image on PASCAL VOC$\rightarrow$Clipart1k is shown in Fig.~\ref{fig:sample}(d). We then feed the target samples and the translated source-like samples into the teacher model, and evaluate their detection performance, respectively. 
   
   The average precisions~(APs) of the teacher models for two versions of samples on different datasets are plotted in Fig.~\ref{fig:test_result}. We observe that the APs of teacher models on source-like samples generally outperform their APs on target samples. 
   This clearly confirms our analysis that the teacher model is biased towards the source domain.

	\subsection{Healing the Model Bias in Mean Teacher}
	\subsubsection{Healing the Teacher Model Bias}
	\label{sec:source_like}
\begin{figure*}[t]
      \centering
        \includegraphics[width=0.8\linewidth]{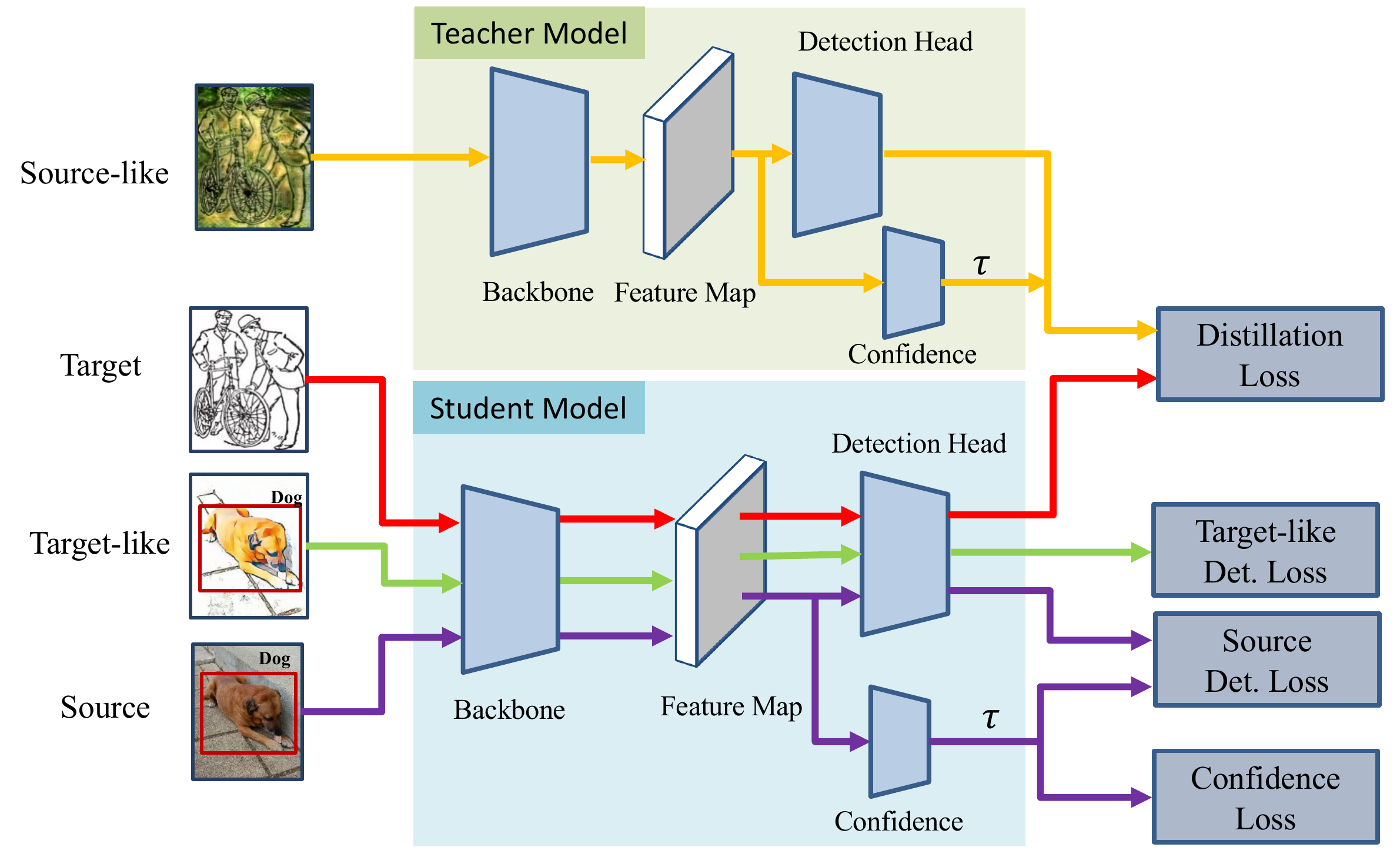}
      \caption{An overview of our proposed Unbiased Mean Teacher. In each mini-batch, four types of images are used: source-like images, target images, target-like images, and source images. The source images with annotations are used to optimize the object detection loss of the student model under the supervision of interpolated ground truth in Eq.~(\ref{eqn:interpolation}) (\textit{i.e.}, Source Det. Loss); the source-like images and target images are respectively fed into the teacher and student network to perform the cross-domain distillation (\textit{i.e.}, Distillation Loss); the target-like images are used as additional training samples to train the student model (\textit{i.e.}, Target-like Det. Loss); The confidence branch adopt proposal feature and predict a confidence score with the log penalty in Eq.~(\ref{eqn:loss_confidence}) (\textit{i.e.}, Confidence Loss). These training routes are performed jointly in an end-to-end manner.}
      \label{fig:framework}
      \vspace{-2mm}
   \end{figure*}

   Motivated by the above observation, we propose to remold the mean teacher model by pixel-level adaptation. Instead of using only target samples for distillation, we perform a \emph{cross-domain distillation} by using paired images $(\I^t, \P^t)$, where $\P^t$ is the source-like version of the target image $\I^t$. As illustrated in~Fig.~\ref{fig:framework}, for each distillation iteration, we feed the source-like image $\P^t$ to the teacher network, and the target image $\I^t$ to the student network. In this way, the teacher network is expected to produce more precise predictions, thus being able to provide better guidance to the student network. Meanwhile, the student network is optimized over the original target samples for the distillation loss, which encourages its favor of target data. Thus, the original distillation loss in Eq.~(\ref{eqn:loss_dist}) is modified as:
   \begin{equation}
    \begin{aligned}
   \mathcal{L}_{dist}(\tilde{\P}^t, \hat{\I}^t) = \cL_{\text{det}}(\cT_B(\tilde{\P}^t), \cT_C(\tilde{\P}^t), \hat{\I}^t),
    \end{aligned}
    \label{eqn:loss_dist_new}
   \end{equation}
   where $\tilde{\P}^t$ is augmented from $\P^t$ with small perturbations.
   \subsubsection{Healing the Student Model Bias}
   \label{sec:target_like}
   As the teacher model is a moving average of the student model, the bias of teacher model essentially comes from the student model. Therefore we also aim to reduce the model bias from the side of student model. Towards this goal, we translate the source images into target-like images. Then the target-like images are used to train the student network, in addition to the supervision from the original source samples. Similarly as in generating source-like images, we utilize CycleGAN to obtain a target-like version image $\P^s$ for each source image $\I^s$ (an example of target-like images is shown in Fig.~\ref{fig:sample}(b)). As the image translation process does not change the ground truth label (\textit{i.e.}, bounding boxes), we use the same label information for target-like images. In this way, the target-like images can encourage the student model to be more favorable of the target domain data, and thus to reduce the bias towards source data. The loss for target-like images can be written as:
   \begin{equation}
   \!\!\!\!\cL_{\text{det}}(\cB^s, \cC^s, \P^s)\!\!=\!\! \cL_{\text{rpn}}(\cB^s;\P^s) + \cL_{\text{roi}}(\cB^s, \cC^s;\P^s),
   \label{eqn:loss_det_new}
   \end{equation}
which has the same form with the loss in Eq.~(\ref{eqn:loss_det}), only replacing $\I^s$ with $\P^s$.
    \subsubsection{Healing the Teaching Process}
    We have illustrated how to alleviate the model bias problem of the teacher and student models by using source-like and target-like images. However, one limitation is that these strategies are static, and cannot be automatically adapted during the training process. Specifically, with the source-like and target-like images, the MT model is expected to be gradually guided to fit the target domain, which means that the source-like images might not be ultimately suitable for the teacher model as the training process goes. 
    
    To this end, we further propose to dynamically adjust the teaching process of the MT model by selecting the most suitable samples for cross-domain distillation. In particular, as discussed in the out-of-distribution detection works~\cite{devries2018learning, hendrycks2016baseline, liang2017enhancing, lee2017training}, the fitness of samples to a model can be measured with their confidences predicted by the model\footnote{Note that the \textit{confidence} here refers to the confidence of a sample being an in-distribution sample, which is different from the prediction confidence (\ie, the maximum category probability) used in other works.}. Therefore, we could employ the target sample confidence scores to automatically select pairs that most fit the \emph{current} teacher model to enhance the teaching process of the MT model during training.
    
    In particular, as shown in Fig.~\ref{fig:framework}, we utilize a side-branch to predict the model confidence for each proposal as inspired by~\cite{devries2018learning}. During the training process, on one hand, we encourage the confidence score to be maximized for all labeled training samples. Let us denote by $\tau_j$ as the confidence score for the $j$-th proposal $B_j$, a log penalty is used to train the confidence prediction branch:
    \begin{equation}
    \cL_{\tau} = \sum_{j}-log(\tau_j).
    \label{eqn:loss_confidence}
    \end{equation}
    On the other hand, the confidence score is used to adjust classification loss through interpolating between the model prediction and the ground-truth label. Formally, we denote $\p_j = [p_j^1, \ldots, p_j^C]^\top$ as the predicted class probability for $B_j$ and $\textbf{y}_j$ as its one-hot representation of the category label. We give the hints through interpolating between the model prediction $\p_j$ and the category distribution $\y_j$. The interpolation degree is indicated by the confidence score $\tau_j$:
    \begin{equation}
    \textbf{p}^{\prime}_j =\tau_j \cdot \textbf{p}_j + (1 - \tau_j) \cdot \textbf{y}_j,
    \label{eqn:interpolation}
    \end{equation}
    in which $\tau_j$ is clipped in the range of $[0, 1]$ if $\tau_j > 1$.
    
    During training, we replace the original one-hot target category probability with the above soft label $\p^{\prime}$. Therefore, Eq.~(\ref{eqn:loss_det}) can be updated as follows:
    \begin{equation}
   \cL_{\text{det}}(\cB^s, \cP, \I^s) = \cL_{\text{rpn}}(\cB^s;\I^s) + \cL_{\text{roi}}(\cB^s, \cP; \I^s),
   \label{eqn:loss_det_interpolation}
   \end{equation}
    where $\cP = \{\p'_j|_{j=1}^M\}$. 
    
    To enhance the teaching process, we utilize the confidence branch to predict the confidence score for source-like image $\tilde{\P}^t$ in the teacher model. A high confidence score indicates the source-like sample is more likely an in-distribution sample for the current model, and is beneficial to be used for the teaching process in the MT model. Therefore, we updated Eq.~(\ref{eqn:threshold}) as follows:
    \begin{equation}
    \begin{aligned}
   (\hat{\cT}_{B}, \hat{\cT}_{C})=\left\{(B_{j}, c_j) | \sqrt{\tau_j \cdot p_{j}}> T, \forall j \right \},
    \end{aligned}
    \label{eqn:threshold_new}
   \end{equation}
   Accordingly, the cross domain distillation loss in Eq.~(\ref{eqn:loss_dist_new}) is updated with the above defined $(\hat{\cT}_{B}, \hat{\cT}_{C})$ as follows:
      \begin{equation}
    \begin{aligned}
   \hat{\mathcal{L}}_{dist}(\tilde{\P}^t, \hat{\I}^t) = \cL_{\text{det}}(\hat{\cT}_B(\tilde{\P}^t), \hat{\cT}_C(\tilde{\P}^t), \hat{\I}^t).
    \end{aligned}
    \label{eqn:loss_dist_new2}
   \end{equation}
\subsubsection{Overall Model}
   We illustrate the overall architecture of our Unbiased Mean Teacher in Fig.~\ref{fig:framework}. The source-like and target-like data are generated offline. Then the model is trained jointly by optimizing all losses in an end-to-end manner. The overall training objective can be written as:
      \begin{equation}
   \begin{split}
   \mathcal{L} = &   \cL_{\text{det}}(\cB^s, \cP, \I^s) +    \cL_{\text{det}}(\cB^s, \cC^s, \P^s) \\ 
   & + \lambda\cdot\hat{\mathcal{L}}_{dist}(\tilde{\P}^t, \hat{\I}^t) + \gamma\cdot\mathcal{L}_{\tau},
   \end{split}
   \label{eq:objective}
   \end{equation} 
   where the loss terms are respectively the detection loss on source samples defined in Eq.~(\ref{eqn:loss_det_interpolation}), the detection loss on target-like samples defined in Eq.~(\ref{eqn:loss_det_new}), the cross-domain distillation loss defined in Eq.~(\ref{eqn:loss_dist_new2}), and the confidence loss defined in~Eq.~(\ref{eqn:loss_confidence}), and $\lambda$ and $\gamma$ are the trade-off parameters. 

	\section{Experiments}
	\label{sec:experiment}
    
    To validate the effectiveness of our approach, we compare with state-of-the-art methods for cross-domain object detection on benchmark datasets with three different types of domain shifts, including 1) real images to artistic images, 2) normal weather to adverse weather, 3) synthetic images to real images. 
   
   As a common practice, we adopt the protocol of unsupervised cross-domain object detection in~\cite{chen2018domain}. Full annotations including the bounding boxes and the corresponding category labels of objects are available for the source domain training data, while the target domain only contains unlabeled images. Moreover, we can access only the unlabeled train set in the target domain, while the target domain test set is strictly held out during the training phase.

   \noindent\textbf{Implementation Details:}
   Following~\cite{chen2018domain}, we take the Faster RCNN~\cite{ren2015faster} model as the base object detection model for our Unbiased Mean Teacher approach. The ResNet-101~\cite{he2016deep} or VGG16~\cite{simonyan2014very} model pre-trained on ImageNet~\cite{krizhevsky2012imagenet} is used as the backbone for the Faster RCNN model. Following the implementation of Faster RCNN with ROI-alignment~\cite{jjfaster2rcnn,he2017mask}, we rescale all images by setting the shorter side of the image to 600 while keeping the image aspect ratios. 
   
   For the mean teacher and our model, unless otherwise stated, we set the trade-off parameter $\lambda=0.01$ and $\gamma=0.1$ for all the experiments. We set the confidence threshold $T = 0.8$ in all our experiments. 
   We train the student network with a learning rate of $0.001$ for the first 50k iterations and schedule linear decay for the learning rate of $0.0001$ for the next 30k iterations. Each batch consists of four image samples: source, target, source-like, and target-like. The weight smooth coefficient parameter $\alpha$ of the exponential moving average for the teacher model is set to $0.99$. Other experimental hyper-parameters settings in our model follow the setup in~\cite{ren2015faster}. 
   
   To understand the individual impact of the proposed components, we include several special versions of our UMT model for ablation study as follows: 1) UMT$_{S}$ is the simple mean teacher model by optimizing the loss in Eq.~(\ref{eqn:loss_mt}); 2) UMT$_{SC}$ is the mean teacher model with our cross-domain distillation strategy as described in Section~\ref{sec:source_like}; and 3) UMT$_{SCA}$ is the mean teacher model with both our cross-domain distillation strategy in Section~\ref{sec:source_like} and using the target-like images to augment the training set for the student model as described in Section~\ref{sec:target_like}.
	
	\subsection{Real to Artistic Adaptation}
	\label{sec:real_to_art}

	\begin{table*}[t]
    \small
    \caption{The average precision (AP, in \%) on all classes from different methods for cross-domain object detection on the Clipart1k test set for \textbf{PASCAL VOC$\rightarrow$Clipart1k} adaptation.
    }
    
    \centering
    \resizebox{\textwidth}{!}{  
      \begin{tabular}{c|c@{ }c@{ }c@{ }c@{ }c@{ }c@{ }c@{ }c@{ }c@{ }c@{ }c@{ }c@{ }c@{ }c@{ }c@{ }c@{ }c@{ }c@{ }c@{ }c|c}
        \hline
        Method &aero&bcycle&bird&boat&bottle&bus&car&cat&chair&cow&table&dog&hrs&bike&prsn&plnt&sheep&sofa&train&tv&MAP\\ \hline \hline
		Source Only &26.0&45.9&23.2&22.1&20.1&51.7&29.8&9.4&34.6&13.6&30.1&0.9&33.7&50.0&37.2&46.2&18.9&6.7&34.1&20.5&27.7\\ \hline
        SCL~\cite{shen2019scl} &\textbf{44.7}&50.0&33.6&27.4&42.2&55.6&38.3&\textbf{19.2}&37.9&\textbf{69.0}&\textbf{30.1}&26.3&34.4&67.3&61.0&47.9&21.4&26.3 &50.1&47.3&41.5\\ \hline
        SWDA~\cite{saito2019strong} &26.2&48.5&32.6&33.7&38.5&54.3&37.1&18.6&34.8&58.3&17.0&12.5&33.8&65.5&61.6&52.0&9.3&24.9 &\textbf{54.1}&49.1&38.1\\ \hline
        DM~\cite{kim2019diversify}& 25.8&63.2&24.5&\textbf{42.4}&\textbf{47.9}&43.1&37.5&9.1&47.0&46.7&26.8&24.9&48.1&\textbf{78.7}&63.0&45.0&21.3&\textbf{36.1} &52.3&\textbf{53.4}&41.8\\ \hline
        CRDA~\cite{xu2020exploring} &28.7 & 55.3& 31.8 &26.0& 40.1& \textbf{63.6} & 36.6& 9.4& 38.7& 49.3& 17.6& 14.1& 33.3& 74.3&61.3& 46.3& 22.3& 24.3& 49.1& 44.3 &38.3 \\ \hline
        HTCN~\cite{chen2020harmonizing} &33.6 &58.9& \textbf{34.0}& 23.4& 45.6& 57.0& 39.8& 12.0& 39.7& 51.3& 21.1& 20.1& 39.1& 72.8& 63.0& 43.1& 19.3& 30.1& 50.2& 51.8& 40.3 \\ \hline
        \multicolumn{1}{c|}{UMT$_S$} &30.9&51.8&27.2&28.0&31.4&59.0&34.2&10.0&35.1&19.6&15.8&9.3&41.6&54.4&52.6&40.3&22.7&28.8&37.8&41.4& 33.6\\ \cline{2-22} 
        \multicolumn{1}{c|}{UMT$_{SC}$}&40.1&\textbf{69.3}&26.8&29.0&24.9&39.4&42.7&8.6&39.8&63.0&14.9&18.8&43.6&66.1&63.0&40.7&\textbf{31.7}&8.7&27.5&53.0&37.6     \\ \cline{2-22} 
        \multicolumn{1}{c|}{UMT$_{SCA}$}&39.5&60.0&30.5&39.7&37.5&56.0&42.7&11.1&\textbf{49.6}&59.5&21.0&29.2&\textbf{49.5}&71.9&66.4&48.0&21.2&13.5&38.8&50.4&41.8     \\ \cline{2-22} 
        \multicolumn{1}{c|}{UMT} &39.6 & 59.1 & 32.4 & 35.0 & 45.1 & 61.9 & \textbf{48.4} & 7.5 & 46.0 & 67.6 & 21.4 & \textbf{29.5} & 48.2 & 75.9 & \textbf{70.5} & \textbf{56.7} & 25.9 & 28.9 & 39.4 & 43.6 & \textbf{44.1} \\ \hline
        Oracle &33.3 & 47.6 & 43.1 & 38.0 & 24.5 & 82.0 & 57.4 & 22.9 & 48.4 & 49.2 & 37.9 & 46.4 & 41.1 & 54.0 & 73.7 & 39.5 & 36.7 & 19.1 & 53.2 & 52.9 & 45.0 \\ \hline
        
      \end{tabular}
      }
    \label{tab:voc_clip}
    
  \end{table*}
    
	\textbf{Datasets}: In this experiment, we test our model with domain shift between the real image domain and the artistic image domain. Following~\cite{saito2019strong,shen2019scl}, we combine the PASCAL VOC 2007 and PASCAL VOC 2012 datasets as the source domain, and use the Clipart1k and Watercolor2k datasets as target domains, respectively.
   The Clipart1k dataset contains $1,000$ images from the same 20 classes as the PASCAL VOC dataset, which is split equally into a training set and a test set, containing $500$ images each. We use the training set as the unlabeled target domain samples for domain adaptation in the training phase, and the test set is held out for evaluation. The Watercolor2k consists of $2,000$ images from $6$ classes in common with the PASCAL VOC dataset. Similarly, we use $1,000$ images as the target unlabeled training data for training models, and the remaining $1,000$ images are held out for testing. 
   
   We include the results from state-of-the-art methods DM~\cite{kim2019diversify} , SWDA~\cite{saito2019strong}, HTCN~\cite{chen2020harmonizing}, CRDA~\cite{xu2020exploring} and SCL~\cite{shen2019scl} for comparison. Besides, we report the oracle result by training a Faster RCNN model using the same images with target domain but with the ground truth annotations, which can be viewed as a reference for the upper bound adaptation performance. All methods are built on the Faster RCNN model, where ImageNet pre-trained ResNet-101~\cite{he2016deep} is used as the backbone network. 
   
   \begin{table}[t]
      \small
      \caption{The average precision (AP, in \%) on all classes from different methods for cross-domain object detection on the Watercolor2k test set for \textbf{PASCAL VOC$\rightarrow$Watercolor2k} adaptation.      
      }
      \centering
      \resizebox{.45\textwidth}{!}{
         \begin{tabular}{c|c@{ }c@{ }c@{ }c@{ }c@{ }c|c}
            \hline
            Method & bike & bird & car & cat & dog & person & MAP \\ \hline \hline
            Source Only & 74.3 & 49.0 & 35.2 & 33.9 & 25.3 & 60.6 & 46.4 \\ \hline 
            SCL~\cite{shen2019scl}&82.2 & 55.1 & 51.8 & 39.6 &38.4 & 64.0 & 55.2\\ \hline 
            DM~\cite{kim2019diversify}&- & - & - & - & - & - & 52.0\\ \hline
            SWDA~\cite{saito2019strong}&82.3 & \textbf{55.9} & 46.5 & 32.7 & 35.5 & 66.7 & 53.3\\ \hline 
            \multicolumn{1}{c|}{UMT$_{S}$} &76.2&53.4&46.2&39.3&34.9&\textbf{71.5}&53.6 \\ \cline{2-8} 
            \multicolumn{1}{c|}{UMT$_{SC}$} &79.7&49.5&50.1&\textbf{45.5}&30.6&69.8&54.2     \\ \cline{2-8}
            \multicolumn{1}{c|}{UMT$_{SCA}$} &86.6&51.3&\textbf{52.6}&42.1&33.5&67.5&55.6     \\ \cline{2-8} 
            \multicolumn{1}{c|}{UMT} & \textbf{88.2} & 55.3 & 51.7 & 39.8 & \textbf{43.6} & 69.9 & \textbf{58.1}   \\ \hline
            oracle & 49.8&50.6&40.2&38.9&53.3&69.4&50.4 \\ \hline
         \end{tabular}
         }
      \label{tab:voc2water}
      \vspace{-4mm}
   \end{table}

	\textbf{Results}: We report the average precision (AP) of each class as well as the mean AP over all classes in Table~\ref{tab:voc_clip} and Table~\ref{tab:voc2water} for object detection on the Clipart1k and Watercolor2k datasets, respectively.  
   
   We take the Clipart1k dataset as an example to explain the experimental results. In particular, the simple MT model UMT$_{S}$ obtains a mean AP of $33.6\%$, which outperforms the result of $27.7\%$ from the source only baseline. This proves that the mean teacher model could help to improve the robustness of object detection model against data variance considerably. However, the improvement is not as significant as other state-of-the-art methods like CRDA, HTCN, and SCL, possibly due to the model bias problem as analyzed in Section~\ref{sec:model_bias}. By using the cross-domain distillation the result is boosted to $37.6\%$ (\textit{i.e.}, UMT$_{SC}$), which is further improved to $41.8\%$ (\textit{i.e.}, UMT$_{SCA}$) by additionally using the target-like augmentation strategy. Note that the result of UMT$_{SC}$ is already on par with the state-of-the-art result on this dataset (\textit{i.e.}, DM), which clearly demonstrates the effectiveness of our strategies for handling the model biases in mean teacher. By dynamically adjusting the teaching process of the MT model by selecting the most suitable samples for cross-domain distillation, our final UMT model reaches $44.1\%$, which gives the new state-of-the-art performance for cross-domain object detection on the Clipart1k dataset. We have similar observations for the Watercolor2k dataset. 
	
	\subsection{Adaptation in Inverse Weather}
	\label{sec:adverse_weather}
   
   \begin{table}[ht]
   \small
      \caption{The average precision (AP, in \%) on all classes from different methods for cross-domain object detection on the validation set of Foggy Cityscapes for \textbf{Cityscapes$\rightarrow$Foggy Cityscapes }
      adaptation. 
      }
      \vspace{-2mm}
      \centering
      \resizebox{\linewidth}{!}{
      \setlength{\tabcolsep}{1pt}
         \begin{tabular}{c|c c c c c c c c|c}
            \hline
            Method & bus & bicycle & car & mcycle & person & rider & train & truck & MAP \\ \hline \hline
            Source Only& 24.7 & 29.0 & 27.2 & 16.4 & 24.3 & 31.5 & 9.1 & 12.1 &21.8 \\ \hline
            
            PF~\cite{shan2019pixel}&- &- &- &- &- &- &-&- &28.9 \\ \hline
            WD~\cite{xu2019wasserstein}&39.9  &34.4 &44.2 &25.4 &30.2 &42.0 &26.5 &22.2 &33.1 \\ \hline
            SCL~\cite{shen2019scl}&41.8  &36.2 &44.8 &33.6 &31.6 &44.0 &40.7 &30.4&37.9 \\ \hline
            DA-Faster~\cite{chen2018domain}&35.3 &27.1& 40.5 &20.0 &25.0 &31.0 &20.2 &22.1 &27.6 \\ \hline
            SCDA~\cite{zhu2019adapting}&39.0 &33.6 &48.5 &28.0 &33.5 & 38.0& 23.3&26.5&33.8 \\ \hline
            DM~\cite{kim2019diversify}&38.4  &32.2 &44.3 &28.4 &30.8 &40.5 &34.5 &27.2 &34.6 \\ \hline
            MAF~\cite{he2019multi}&39.9& 33.9 &43.9 & 29.2&28.2 &39.5 &33.3 &23.8 &34.0 \\ \hline

            MTOR~\cite{cai2019exploring}&38.6  &35.6 &44.0 &28.3 &30.6 &41.4 &40.6 &21.9 &35.1 \\ \hline
            SWDA~\cite{saito2019strong}&36.2 &35.3& 43.5 &30.0 &29.9 &42.3 &32.6 &24.5 &34.3 \\ \hline 
            CRDA~\cite{xu2020exploring} & 45.1 & 34.6 &49.2&30.3&32.9&43.8&36.4&27.2 &37.4 \\ \hline
            HTCN~\cite{chen2020harmonizing} & 47.4 & 37.1 & 47.9 & \textbf{32.3} &33.2 &\textbf{47.5} &40.9 &31.6 &39.8 \\ \hline  
            iFAN~\cite{zhuang2020ifan} & 45.5 & 33.0& 48.5& 22.8& 32.6& 40.0& 31.7& 27.9& 35.3 \\ \hline
            CDN~\cite{su2020adapting} & 42.5& 36.5& \textbf{50.9}& 30.8& \textbf{35.8}& 45.7& 29.8& 30.1& 36.6 \\ \hline
            \multicolumn{1}{c|}{UMT$_S$}&30.1&31.3&36.1&22.4&27.9&38.2&20.2&21.5&28.5 \\ \cline{2-10} 
            \multicolumn{1}{c|}{UMT$_{SC}$}&43.4 &38.0 & 50.6& 33.7& 33.4& 45.9& 36.4& 31.9& 39.2 \\ \cline{2-10} 
            \multicolumn{1}{c|}{UMT$_{SCA}$}&48.2&\textbf{38.9}&49.8&33.0&33.8&47.3&42.1&30.0&40.4 \\ \cline{2-10}
            \multicolumn{1}{c|}{UMT} &\textbf{56.5}& 37.3& 48.6 &30.4 & 33.0& 46.7 & \textbf{46.8} & \textbf{34.1} & \textbf{41.7}   \\ \hline
            Oracle &50.0 & 36.2 & 49.7 & 34.7 & 33.2 & 45.9 & 37.4 & 35.6 &40.3 \\ \hline
         \end{tabular}
         }
      \label{tab:city2foggy}
      \vspace{-4mm}
   \end{table}
    
	\textbf{Datasets}: In this experiment, we follow the setting in~\cite{chen2018domain}. The training set of the Cityscapes dataset is used as the source domain, and the Foggy Cityscapes dataset~\cite{sakaridis2018semantic} is used as the target domain. 
    The Cityscapes dataset is collected from the urban street scene captured in 50 cities. The dataset contains $2,975$ images in the train set and $500$ images in the validation set. The Foggy Cityscapes is a synthetic foggy scene dataset rendered using the images and depth maps from Cityscapes, which therefore has the same data split as the Cityscapes dataset, \textit{i.e.} a training set of $2,975$ images a validation set of $500$ images. We take labeled Cityscapes train set images and unlabeled Foggy Cityscapes train set images in our experiment, and report the evaluated results on the validation set of Foggy Cityscapes. Although there exists a one-to-one correspondence between images in Cityscapes and Foggy Cityscapes datasets, we do not leverage such information in unsupervised domain adaptation.

    Besides the baselines compared previously, we further include DA-Faster~\cite{chen2018domain}, PF~\cite{shan2019pixel}, SCDA~\cite{zhu2019adapting}, MAF~\cite{he2019multi}, WD~\cite{xu2019wasserstein}, MTOR~\cite{cai2019exploring}, iFAN~\cite{zhuang2020ifan} and CDN~\cite{su2020adapting} for comparison. The setup for special cases of our UMT approach and the oracle method is the same as those in the previous experiments. All methods are built on Faster RCNN, where the VGG16~\cite{simonyan2014very} pre-trained on ImageNet~\cite{krizhevsky2012imagenet} is used as the backbone.
	
	\textbf{Results}: Object detection in foggy scene images is extremely challenging due to low visibility. The current best state-of-the-art result on this dataset is $39.8\%$ from the recent work HTCN~\cite{chen2020harmonizing}. However, the special case of our UMT$_{SC}$ model which using mean teacher with cross-domain distillation already approaches HTCN with a mean AP of $39.2\%$. This again validates the effectiveness of the cross-domain distillation strategy for improving the mean teacher model for cross-domain object detection. By healing the student model bias and teaching process, our final UMT model shows improved performance compared to HTCN and reaches a mean AP of $41.7\%$. Interestingly, this result exceeds the oracle result on this dataset, showing that the clear weather images with high visibility are useful for boosting the limitation of the object detection in the adverse foggy weather with low visibility, without requiring any annotations on those low visibility images.

	\subsection{Synthetic-to-Real Adaptation}
	\vspace{-2mm}
	\begin{table}[t]
	 \small
      \caption{The average precision (AP, in \%) of different methods for cross-domain object detection on the validation set of Cityscapes for \textbf{SIM10K$\rightarrow$Cityscapes}
      adaptation. 
      }
      \centering
      \resizebox{0.2\textwidth}{!}{
      \setlength{\tabcolsep}{1pt}
         \begin{tabular}{c | c }
            \hline
            Method & AP on \textit{car} \\ \hline \hline
            Source Only & 34.3 \\ \hline
            WD~\cite{xu2019wasserstein} & 40.6 \\ \hline  
            SCL~\cite{shen2019scl} & 42.6 \\ \hline
            DA-Faster~\cite{chen2018domain} & 38.97 \\ \hline
            SCDA~\cite{zhu2019adapting} & 43.0 \\ \hline
            SWDA~\cite{saito2019strong} & 40.1 \\ \hline
            MAF~\cite{he2019multi} & 41.1 \\ \hline
            HTCN~\cite{chen2020harmonizing} & 42.5 \\ \hline
            \multicolumn{1}{c|}{UMT$_S$} & 40.8 \\ \cline{2-2} 
            \multicolumn{1}{c|}{UMT$_{SC}$} & 42.0 \\ \cline{2-2} 
            \multicolumn{1}{c|}{UMT$_{SCA}$} & 42.6 \\ \cline{2-2} 
            \multicolumn{1}{c|}{UMT} & \textbf{43.1} \\ \hline
            Oracle & 53.0 \\ \hline
         \end{tabular}
         }
      \label{tab:sim10k2city}
      \vspace{-4mm}
   \end{table}
    
	\textbf{Datasets}: Following~\cite{chen2018domain}, the SIM10K dataset~\cite{johnson2016driving} is used as the source domain, and the Cityscapes dataset is used as the target domain. The SIM10K dataset contains $10,000$ images of the computer-rendered driving scene from the Grand Theft Auto (GTAV) game. The training set of Cityscapes is used as target training samples, and the validation set is used for evaluation. All methods are built on Faster RCNN, where the VGG16~\cite{simonyan2014very} pre-trained on ImageNet~\cite{krizhevsky2012imagenet} is used as the backbone.
	
	\textbf{Results}: The AP on detecting cars for different approaches are reported in Table~\ref{tab:sim10k2city}. Similarly, as in the previous experiments, we observe that our UMT approach gradually improves the MT model by addressing its model bias with different strategies. Our final model also achieves the new state-of-the-art AP of $43.1\%$ on this dataset using VGG16~\cite{simonyan2014very} as the backbone, which again demonstrates the effectiveness of our proposed approach.
	
	\section{Conclusion}
	In this work, we provide a novel perspective to study the cross-domain object detection problem by exploiting the observation that a detection model can be easily biased towards source images. For that, we present a new method, named Unbiased Mean Teacher (UMT), by designing three highly effective strategies to remedy the model bias. In particular, we firstly introduce cross-domain distillation to maximally exploit the expertise of the teacher model. Then, we further augment the training samples for the student model through pixel-level adaptation to reduce its model bias. Lastly, we employ an out-of-distribution estimation strategy to select samples that most fit the current model to enhance the cross-domain distillation process. Extensive experiments are conducted on multiple benchmark datasets, and the results clearly show that our UMT surpasses the existing state-of-the-art models by relatively large margins.
    \noindent \textbf{Acknowledgement:} This work is partially supported by the Major Project for New Generation of AI under Grant No. 2018AAA0100400 and China Postdoctoral Science Foundation(NO.2019TQ0051).

{\small
\bibliographystyle{ieee_fullname}
\bibliography{egbib}

\begin{thebibliography}{10}\itemsep=-1pt

\bibitem{cai2019exploring}
Qi Cai, Yingwei Pan, Chong-Wah Ngo, Xinmei Tian, Lingyu Duan, and Ting Yao.
\newblock Exploring object relation in mean teacher for cross-domain detection.
\newblock In {\em Proceedings of the IEEE/CVF Conference on Computer Vision and
  Pattern Recognition}, pages 11457--11466, 2019.

\bibitem{cai2018cascade}
Zhaowei Cai and Nuno Vasconcelos.
\newblock {Cascade R-CNN}: Delving into high quality object detection.
\newblock In {\em Proceedings of the IEEE/CVF Conference on Computer Vision and
  Pattern Recognition}, pages 6154--6162, 2018.

\bibitem{cariucci2017autodial}
Fabio~Maria Cariucci, Lorenzo Porzi, Barbara Caputo, Elisa Ricci, and Samuel
  Rota~Bulo.
\newblock {AutoDIAL}: Automatic domain alignment layers.
\newblock In {\em Proceedings of the IEEE/CVF International Conference on
  Computer Vision}, pages 5067--5075, 2017.

\bibitem{chen2020harmonizing}
Chaoqi Chen, Zebiao Zheng, Xinghao Ding, Yue Huang, and Qi Dou.
\newblock Harmonizing transferability and discriminability for adapting object
  detectors.
\newblock In {\em Proceedings of the IEEE/CVF Conference on Computer Vision and
  Pattern Recognition}, pages 8869--8878, 2020.

\bibitem{chen2019learning}
Yuhua Chen, Wen Li, Xiaoran Chen, and Luc~Van Gool.
\newblock Learning semantic segmentation from synthetic data: A geometrically
  guided input-output adaptation approach.
\newblock In {\em Proceedings of the IEEE/CVF Conference on Computer Vision and
  Pattern Recognition}, pages 1841--1850, 2019.

\bibitem{chen2018domain}
Yuhua Chen, Wen Li, Christos Sakaridis, Dengxin Dai, and Luc Van~Gool.
\newblock Domain adaptive faster r-cnn for object detection in the wild.
\newblock In {\em Proceedings of the IEEE/CVF Conference on Computer Vision and
  Pattern Recognition}, pages 3339--3348, 2018.

\bibitem{devries2018learning}
Terrance DeVries and Graham~W Taylor.
\newblock Learning confidence for out-of-distribution detection in neural
  networks.
\newblock {\em arXiv preprint arXiv:1802.04865}, 2018.

\bibitem{everingham2010pascal}
Mark Everingham, Luc Van~Gool, Christopher~KI Williams, John Winn, and Andrew
  Zisserman.
\newblock The pascal visual object classes (voc) challenge.
\newblock {\em International Journal of Computer Vision}, 88(2):303--338, 2010.

\bibitem{french2017self}
Geoffrey French, Michal Mackiewicz, and Mark Fisher.
\newblock Self-ensembling for visual domain adaptation.
\newblock {\em arXiv preprint arXiv:1706.05208}, 2017.

\bibitem{ganin2014unsupervised}
Yaroslav Ganin and Victor Lempitsky.
\newblock Unsupervised domain adaptation by backpropagation.
\newblock In {\em International Conference on Machine Learning}, pages
  1180--1189, 2015.

\bibitem{ganin2016domain}
Yaroslav Ganin, Evgeniya Ustinova, Hana Ajakan, Pascal Germain, Hugo
  Larochelle, Fran{\c{c}}ois Laviolette, Mario Marchand, and Victor Lempitsky.
\newblock Domain-adversarial training of neural networks.
\newblock {\em The Journal of Machine Learning Research}, 17(1):2096--2030,
  2016.

\bibitem{girshick2015fast}
Ross Girshick.
\newblock {Fast R-CNN}.
\newblock In {\em Proceedings of the IEEE/CVF International Conference on
  Computer Vision}, pages 1440--1448, 2015.

\bibitem{girshick2014rich}
Ross Girshick, Jeff Donahue, Trevor Darrell, and Jitendra Malik.
\newblock Rich feature hierarchies for accurate object detection and semantic
  segmentation.
\newblock In {\em Proceedings of the IEEE/CVF Conference on Computer Vision and
  Pattern Recognition}, pages 580--587, 2014.

\bibitem{he2017mask}
Kaiming He, Georgia Gkioxari, Piotr Doll{\'a}r, and Ross Girshick.
\newblock Mask {R-CNN}.
\newblock In {\em Proceedings of the IEEE/CVF International Conference on
  Computer Vision}, pages 2961--2969, 2017.

\bibitem{he2015spatial}
Kaiming He, Xiangyu Zhang, Shaoqing Ren, and Jian Sun.
\newblock Spatial pyramid pooling in deep convolutional networks for visual
  recognition.
\newblock {\em IEEE Transactions on Pattern Analysis and Machine Intelligence},
  37(9):1904--1916, 2015.

\bibitem{he2016deep}
Kaiming He, Xiangyu Zhang, Shaoqing Ren, and Jian Sun.
\newblock Deep residual learning for image recognition.
\newblock In {\em Proceedings of the IEEE/CVF Conference on Computer Vision and
  Pattern Recognition}, pages 770--778, 2016.

\bibitem{he2019multi}
Zhenwei He and Lei Zhang.
\newblock Multi-adversarial faster-rcnn for unrestricted object detection.
\newblock In {\em Proceedings of the IEEE/CVF International Conference on
  Computer Vision}, pages 6668--6677, 2019.

\bibitem{hendrycks2016baseline}
Dan Hendrycks and Kevin Gimpel.
\newblock A baseline for detecting misclassified and out-of-distribution
  examples in neural networks.
\newblock {\em arXiv preprint arXiv:1610.02136}, 2016.

\bibitem{inoue2018cross}
Naoto Inoue, Ryosuke Furuta, Toshihiko Yamasaki, and Kiyoharu Aizawa.
\newblock Cross-domain weakly-supervised object detection through progressive
  domain adaptation.
\newblock In {\em Proceedings of the IEEE/CVF Conference on Computer Vision and
  Pattern Recognition}, pages 5001--5009, 2018.

\bibitem{johnson2016driving}
Matthew Johnson-Roberson, Charles Barto, Rounak Mehta, Sharath~Nittur Sridhar,
  Karl Rosaen, and Ram Vasudevan.
\newblock Driving in the matrix: Can virtual worlds replace human-generated
  annotations for real world tasks?
\newblock {\em arXiv preprint arXiv:1610.01983}, 2016.

\bibitem{khodabandeh2019robust}
Mehran Khodabandeh, Arash Vahdat, Mani Ranjbar, and William~G Macready.
\newblock A robust learning approach to domain adaptive object detection.
\newblock In {\em Proceedings of the IEEE/CVF International Conference on
  Computer Vision}, pages 480--490, 2019.

\bibitem{kim2019self}
Seunghyeon Kim, Jaehoon Choi, Taekyung Kim, and Changick Kim.
\newblock Self-training and adversarial background regularization for
  unsupervised domain adaptive one-stage object detection.
\newblock In {\em Proceedings of the IEEE/CVF International Conference on
  Computer Vision}, pages 6092--6101, 2019.

\bibitem{kim2019diversify}
Taekyung Kim, Minki Jeong, Seunghyeon Kim, Seokeon Choi, and Changick Kim.
\newblock Diversify and match: A domain adaptive representation learning
  paradigm for object detection.
\newblock In {\em Proceedings of the IEEE/CVF Conference on Computer Vision and
  Pattern Recognition}, pages 12456--12465, 2019.

\bibitem{krizhevsky2012imagenet}
Alex Krizhevsky, Ilya Sutskever, and Geoffrey~E Hinton.
\newblock Imagenet classification with deep convolutional neural networks.
\newblock In {\em Advances in Neural Information Processing Systems}, pages
  1097--1105, 2012.

\bibitem{lee2017training}
Kimin Lee, Honglak Lee, Kibok Lee, and Jinwoo Shin.
\newblock Training confidence-calibrated classifiers for detecting
  out-of-distribution samples.
\newblock {\em arXiv preprint arXiv:1711.09325}, 2017.

\bibitem{liang2017enhancing}
Shiyu Liang, Yixuan Li, and Rayadurgam Srikant.
\newblock Enhancing the reliability of out-of-distribution image detection in
  neural networks.
\newblock {\em arXiv preprint arXiv:1706.02690}, 2017.

\bibitem{lin2017feature}
Tsung-Yi Lin, Piotr Doll{\'a}r, Ross Girshick, Kaiming He, Bharath Hariharan,
  and Serge Belongie.
\newblock Feature pyramid networks for object detection.
\newblock In {\em Proceedings of the IEEE/CVF Conference on Computer Vision and
  Pattern Recognition}, pages 2117--2125, 2017.

\bibitem{lin2014microsoft}
Tsung-Yi Lin, Michael Maire, Serge Belongie, James Hays, Pietro Perona, Deva
  Ramanan, Piotr Doll{\'a}r, and C~Lawrence Zitnick.
\newblock Microsoft {COCO}: Common objects in context.
\newblock In {\em European Conference on Computer Vision}, pages 740--755,
  2014.

\bibitem{liu2016ssd}
Wei Liu, Dragomir Anguelov, Dumitru Erhan, Christian Szegedy, Scott Reed,
  Cheng-Yang Fu, and Alexander~C Berg.
\newblock {SSD}: Single shot multibox detector.
\newblock In {\em European Conference on Computer Vision}, pages 21--37, 2016.

\bibitem{long2015learning}
Mingsheng Long, Yue Cao, Jianmin Wang, and Michael Jordan.
\newblock Learning transferable features with deep adaptation networks.
\newblock In {\em International Conference on Machine Learning}, pages 97--105,
  2015.

\bibitem{long2016unsupervised}
Mingsheng Long, Han Zhu, Jianmin Wang, and Michael~I Jordan.
\newblock Unsupervised domain adaptation with residual transfer networks.
\newblock In {\em Advances in Neural Information Processing Systems}, pages
  136--144, 2016.

\bibitem{long2017deep}
Mingsheng Long, Han Zhu, Jianmin Wang, and Michael~I Jordan.
\newblock Deep transfer learning with joint adaptation networks.
\newblock In {\em International Conference on Machine Learning}, pages
  2208--2217, 2017.

\bibitem{redmon2016you}
Joseph Redmon, Santosh Divvala, Ross Girshick, and Ali Farhadi.
\newblock You only look once: Unified, real-time object detection.
\newblock In {\em Proceedings of the IEEE/CVF Conference on Computer Vision and
  Pattern Recognition}, pages 779--788, 2016.

\bibitem{redmon2017yolo9000}
Joseph Redmon and Ali Farhadi.
\newblock {YOLO9000}: better, faster, stronger.
\newblock In {\em Proceedings of the IEEE/CVF Conference on Computer Vision and
  Pattern Recognition}, pages 7263--7271, 2017.

\bibitem{ren2015faster}
Shaoqing Ren, Kaiming He, Ross Girshick, and Jian Sun.
\newblock Faster {R-CNN}: towards real-time object detection with region
  proposal networks.
\newblock {\em IEEE Transactions on Pattern Analysis and Machine Intelligence},
  39(6):1137--1149, 2016.

\bibitem{roychowdhury2019automatic}
Aruni RoyChowdhury, Prithvijit Chakrabarty, Ashish Singh, SouYoung Jin, Huaizu
  Jiang, Liangliang Cao, and Erik Learned-Miller.
\newblock Automatic adaptation of object detectors to new domains using
  self-training.
\newblock In {\em Proceedings of the IEEE/CVF Conference on Computer Vision and
  Pattern Recognition}, pages 780--790, 2019.

\bibitem{saito2019strong}
Kuniaki Saito, Yoshitaka Ushiku, Tatsuya Harada, and Kate Saenko.
\newblock Strong-weak distribution alignment for adaptive object detection.
\newblock In {\em Proceedings of the IEEE/CVF Conference on Computer Vision and
  Pattern Recognition}, pages 6956--6965, 2019.

\bibitem{sakaridis2018semantic}
Christos Sakaridis, Dengxin Dai, and Luc Van~Gool.
\newblock Semantic foggy scene understanding with synthetic data.
\newblock {\em International Journal of Computer Vision}, 126(9):973--992,
  2018.

\bibitem{sankaranarayanan2018generate}
Swami Sankaranarayanan, Yogesh Balaji, Carlos~D Castillo, and Rama Chellappa.
\newblock Generate to adapt: Aligning domains using generative adversarial
  networks.
\newblock In {\em Proceedings of the IEEE/CVF Conference on Computer Vision and
  Pattern Recognition}, pages 8503--8512, 2018.

\bibitem{shan2019pixel}
Yuhu Shan, Wen~Feng Lu, and Chee~Meng Chew.
\newblock Pixel and feature level based domain adaptation for object detection
  in autonomous driving.
\newblock {\em Neurocomputing}, 367:31--38, 2019.

\bibitem{shen2017dsod}
Zhiqiang Shen, Zhuang Liu, Jianguo Li, Yu-Gang Jiang, Yurong Chen, and
  Xiangyang Xue.
\newblock {DSOD}: Learning deeply supervised object detectors from scratch.
\newblock In {\em Proceedings of the IEEE/CVF International Conference on
  Computer Vision}, pages 1919--1927, 2017.

\bibitem{shen2019scl}
Zhiqiang Shen, Harsh Maheshwari, Weichen Yao, and Marios Savvides.
\newblock {SCL}: Towards accurate domain adaptive object detection via gradient
  detach based stacked complementary losses.
\newblock {\em arXiv preprint arXiv:1911.02559}, 2019.

\bibitem{simonyan2014very}
Karen Simonyan and Andrew Zisserman.
\newblock Very deep convolutional networks for large-scale image recognition.
\newblock {\em arXiv preprint arXiv:1409.1556}, 2014.

\bibitem{su2020adapting}
Peng Su, Kun Wang, Xingyu Zeng, Shixiang Tang, Dapeng Chen, Di Qiu, and
  Xiaogang Wang.
\newblock Adapting object detectors with conditional domain normalization.
\newblock {\em arXiv preprint arXiv:2003.07071}, 2020.

\bibitem{sun2016deep}
Baochen Sun and Kate Saenko.
\newblock Deep coral: Correlation alignment for deep domain adaptation.
\newblock In {\em European Conference on Computer Vision}, pages 443--450,
  2016.

\bibitem{tarvainen2017mean}
Antti Tarvainen and Harri Valpola.
\newblock Mean teachers are better role models: Weight-averaged consistency
  targets improve semi-supervised deep learning results.
\newblock In {\em Advances in Neural Information Processing Systems}, pages
  1195--1204, 2017.

\bibitem{tzeng2018splat}
Eric Tzeng, Kaylee Burns, Kate Saenko, and Trevor Darrell.
\newblock Splat: Semantic pixel-level adaptation transforms for detection.
\newblock {\em arXiv preprint arXiv:1812.00929}, 2018.

\bibitem{tzeng2017adversarial}
Eric Tzeng, Judy Hoffman, Kate Saenko, and Trevor Darrell.
\newblock Adversarial discriminative domain adaptation.
\newblock In {\em Proceedings of the IEEE/CVF Conference on Computer Vision and
  Pattern Recognition}, pages 7167--7176, 2017.

\bibitem{uijlings2013selective}
Jasper~RR Uijlings, Koen~EA Van De~Sande, Theo Gevers, and Arnold~WM Smeulders.
\newblock Selective search for object recognition.
\newblock {\em International Journal of Computer Vision}, 104(2):154--171,
  2013.

\bibitem{xie2019multi}
Rongchang Xie, Fei Yu, Jiachao Wang, Yizhou Wang, and Li Zhang.
\newblock Multi-level domain adaptive learning for cross-domain detection.
\newblock In {\em Proceedings of the IEEE/CVF International Conference on
  Computer Vision Workshops}, pages 0--0, 2019.

\bibitem{xu2020exploring}
Chang-Dong Xu, Xing-Ran Zhao, Xin Jin, and Xiu-Shen Wei.
\newblock Exploring categorical regularization for domain adaptive object
  detection.
\newblock In {\em Proceedings of the IEEE/CVF Conference on Computer Vision and
  Pattern Recognition}, pages 11724--11733, 2020.

\bibitem{xu2019wasserstein}
Pengcheng Xu, Prudhvi Gurram, Gene Whipps, and Rama Chellappa.
\newblock Wasserstein distance based domain adaptation for object detection.
\newblock {\em arXiv preprint arXiv:1909.08675}, 2019.

\bibitem{jjfaster2rcnn}
Jianwei Yang, Jiasen Lu, Dhruv Batra, and Devi Parikh.
\newblock A faster pytorch implementation of faster r-cnn.
\newblock {\em https://github.com/jwyang/faster-rcnn.pytorch}, 2017.

\bibitem{zhang2018collaborative}
Weichen Zhang, Wanli Ouyang, Wen Li, and Dong Xu.
\newblock Collaborative and adversarial network for unsupervised domain
  adaptation.
\newblock In {\em Proceedings of the IEEE/CVF Conference on Computer Vision and
  Pattern Recognition}, pages 3801--3809, 2018.

\bibitem{zhu2017unpaired}
Jun-Yan Zhu, Taesung Park, Phillip Isola, and Alexei~A Efros.
\newblock Unpaired image-to-image translation using cycle-consistent
  adversarial networks.
\newblock In {\em Proceedings of the IEEE/CVF International Conference on
  Computer Vision}, pages 2223--2232, 2017.

\bibitem{zhu2019adapting}
Xinge Zhu, Jiangmiao Pang, Ceyuan Yang, Jianping Shi, and Dahua Lin.
\newblock Adapting object detectors via selective cross-domain alignment.
\newblock In {\em Proceedings of the IEEE/CVF Conference on Computer Vision and
  Pattern Recognition}, pages 687--696, 2019.

\bibitem{zhuang2020ifan}
Chenfan Zhuang, Xintong Han, Weilin Huang, and Matthew Scott.
\newblock ifan: Image-instance full alignment networks for adaptive object
  detection.
\newblock In {\em Proceedings of the AAAI Conference on Artificial
  Intelligence}, volume~34, pages 13122--13129, 2020.

\end{thebibliography}
}

\clearpage
\begin{appendices}

In this section, we provide detection error analysis, parameter analysis and qualitative results.

\section{Detection Error Analysis} 
To further understand the individual impact of each component of our UMT model, we experiment to study the detection errors occurred in different models. We use \textbf{PASCAL VOC$\rightarrow$Clipart1k} for the analysis. The experiment settings remain the same as in the main paper. The results from the baseline model (Source Only) and our proposed UMT along with its ablated versions UMT$_{S}$, UMT$_{SC}$, UMT$_{SCA}$ are examined in this experiment.

\subsection{Localization Errors}
  Following DA-Faster\cite{chen2018domain}, we analyze accuracies from the top ranked detections for localization errors. We categorize the detection error into three types, based on the overlap (IoU) between prediction and ground truth bounding box: \textbf{Correct}~(IoU $\geq 0.5$), \textbf{Mis-Localized}~(IoU $\in (0.3,0.5)$) and \textbf{Background}~(IoU $\leq 0.3$). 

\begin{figure}[ht]
  \centering
    \includegraphics[width=1\linewidth]{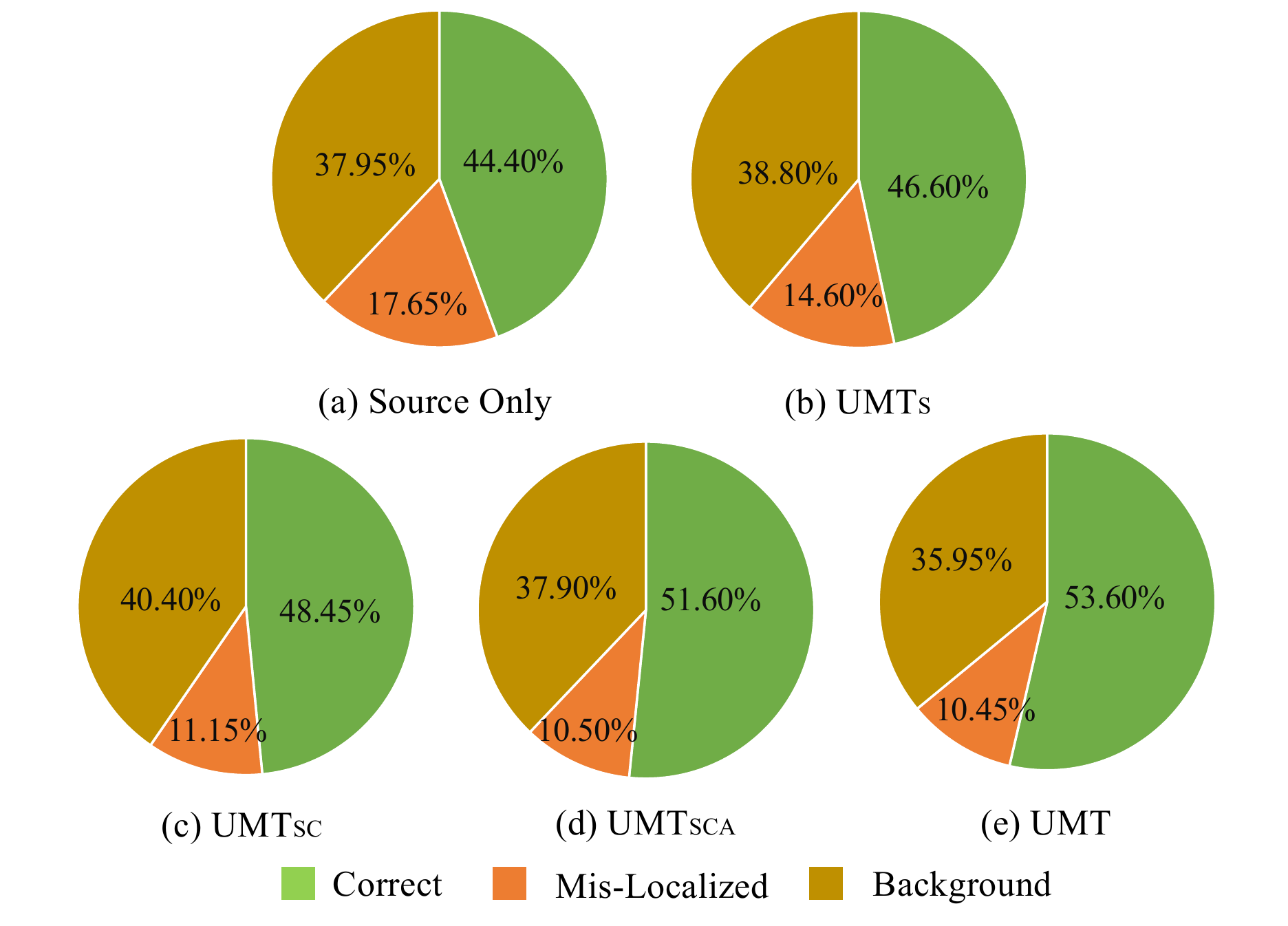}
  \caption{Error analysis on most-confident detections.} 
  \label{fig:loc_error}
\end{figure}

The error distributions are illustrated in Fig.~\ref{fig:loc_error}. We observe that our model can consistently improve the detection accuracy by integrating different strategies for healing the model bias of mean teacher (from $46.60\%$ to $48.45\%$, $51.60\%$, and $53.60\%$). 

With the simple version of the mean teacher model~(\textit{i.e.}, UMT$_{S}$), we observe that detection accuracy is improved compared to the Source Only baseline (from $44.40\%$ to $46.60\%$), while the localization error is also reduced from $17.65\%$ to $14.60\%$. Furthermore, by using the cross-domain distillation strategy, the UMT$_{SC}$ model improves the detection accuracy by a considerable margin (from $46.60\%$ to $48.45\%$), which mostly attributes to the reduction in the Mis-Classified error (from $14.60\%$ to $11.15\%$). This clearly shows the efficacy of using the source-like samples to correct the model bias of the teacher network. Meanwhile, by using the target-like samples to heal the model bias of the student network, the detection accuracy is improved to $51.60\%$. Finally, by employing an out-of-distribution estimation strategy to select samples that most fit the current model to further enhance the cross-domain distillation process, our $UMT$ model achieves an improvement of $9.60\%$ in terms of accuracy compared to the Source Only baseline. This again demonstrates the effectiveness of our proposed model for cross-domain object detection. 

\subsection{Classification Errors on Maximum Overlap Detections}
\begin{figure}[ht]
  \begin{center}
    \includegraphics[width=1\linewidth]{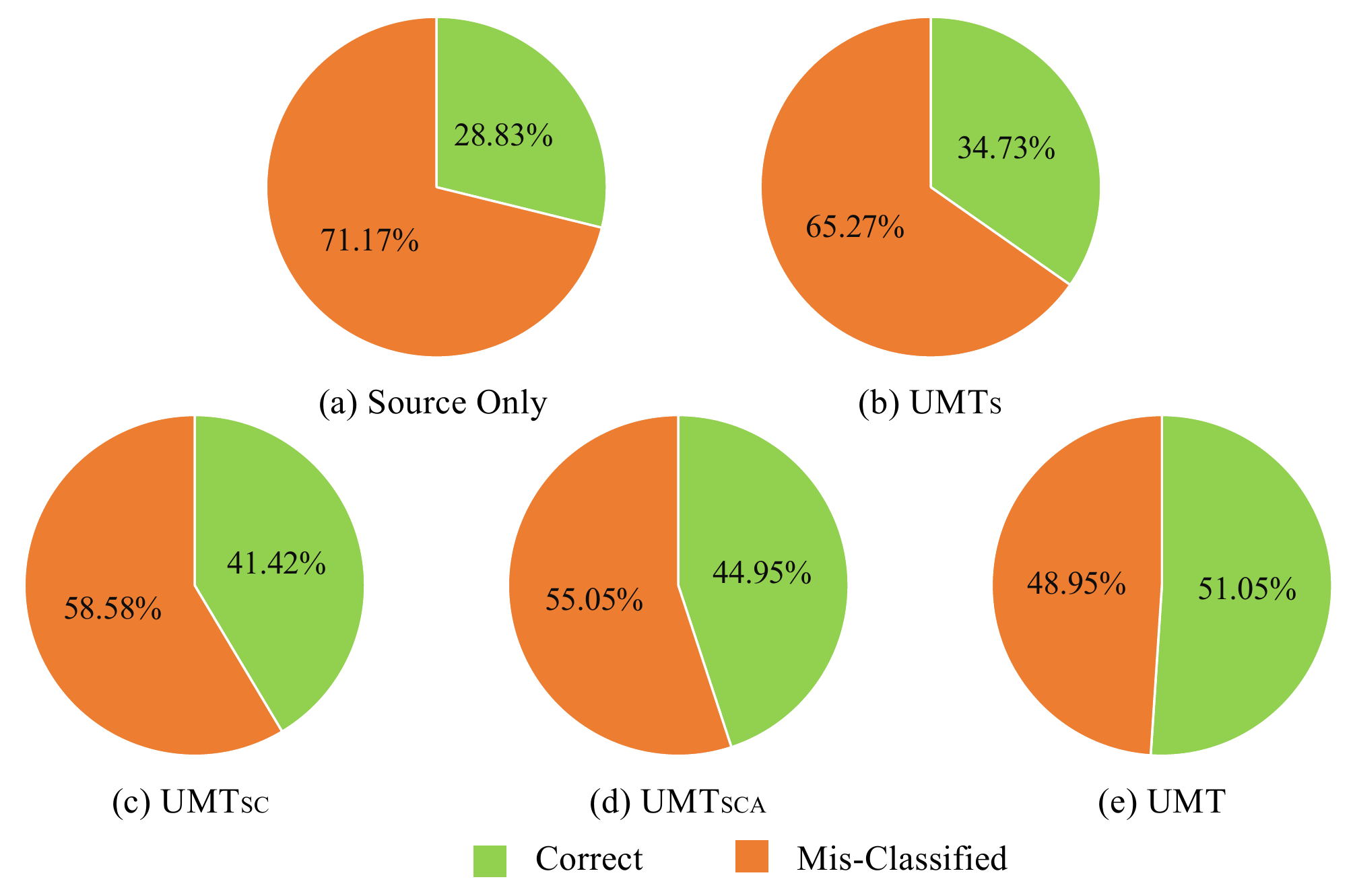}
  \end{center}
  \caption{Classification error analysis on maximum overlap detections.} 
  \label{fig:cls_error}
\end{figure}

We also study the classification error made by each model. For this purpose, we compute the accuracy of the detections which have the highest overlap with the ground truth. As shown in Fig~\ref{fig:cls_error}, though the localization error is mostly solved, there is still a significant amount of classification error for all models. The results suggest that the classification error is relatively more prominent to localization error for cross-domain object detection. Nevertheless, our proposed components can effectively reduce the classification error and improve cross-domain detection performance. Compared with Source Only, our UMT improves the number of correct classified~(green color) and reduces the number of Mis-Classified~(orange color) detections.

\section{Parameter Analysis}
\begin{figure}[t]
  \centering
     \includegraphics[width=1.0\linewidth]{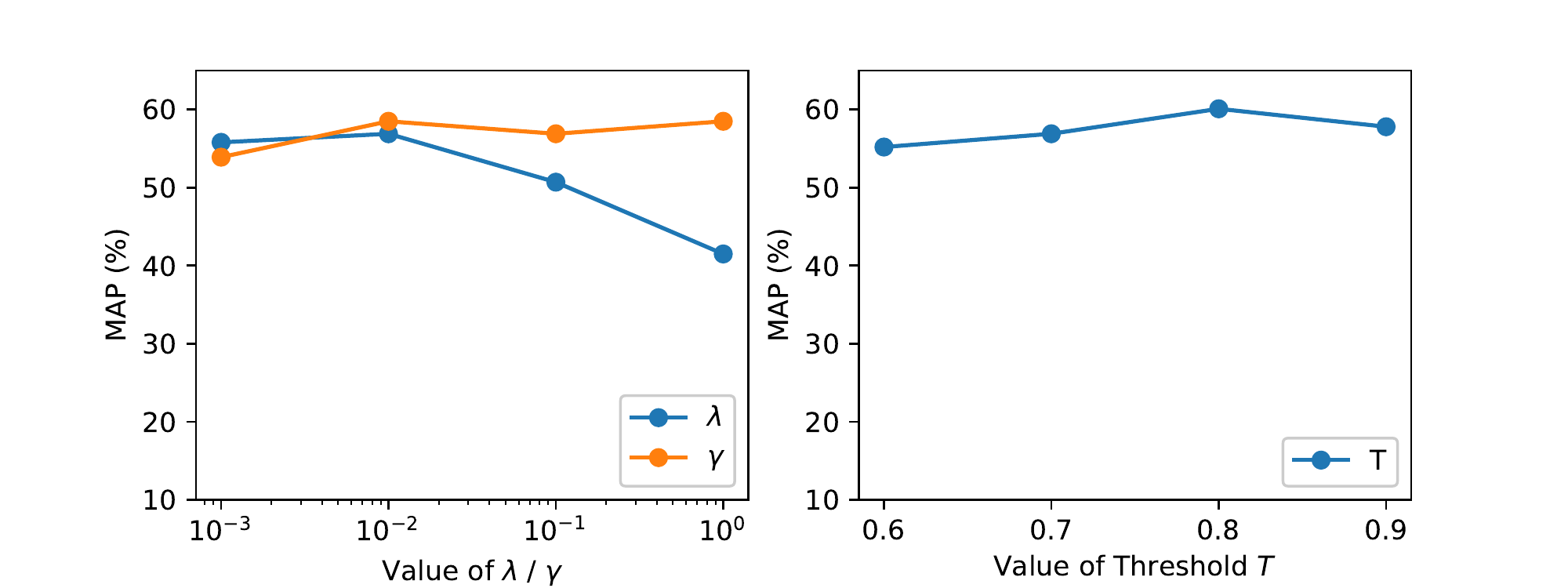}
  \caption{The average precision (AP, in \%) w.r.t. variational values of $\lambda$, $\gamma$ and threshold $T$ in \textbf{PASCAL VOC$\rightarrow$Watercolor2k} adaptation.
  }
  \label{fig:para}
\end{figure}
  
\textbf{Sensitivity of trade-off parameters:} Recall that our UMT method contains three hyper-parameters: the threshold $T$ which determines whether a bounding box is a confident prediction, and the trade-off parameters $\lambda$ and $\gamma$. As stated in the main paper, we empirically set $\lambda$, $\gamma$, and $T$ to $0.01$, $0.1$, and $0.8$ respectively in our experiments. Here we further analyze the effect of parameters by varying the values of $\lambda$, $\gamma$, and $T$, respectively. When varying one parameter, the other two parameters are fixed as their default values. The performance variations \wrt different values of $\lambda$, $\gamma$ and $T$ are shown in Fig~\ref{fig:para}. We can see that the parameters $\gamma$ and $T$ is not sensitive and achieve consistently high detection performance, and the $\lambda$ obtained with smaller values $\lambda=0.001, 0.01$ are generally better than those achieved by larger values. Though our UMT approach generally achieves good performance by setting $\lambda$, $\gamma$, and $T$ to fixed values, we also observe that for a specific scenario like {PASCAL VOC$\rightarrow$Clipart1k}  better results can be obtained by varying the values of parameters.

\begin{figure}[t]
  \centering
     \includegraphics[width=0.9\linewidth]{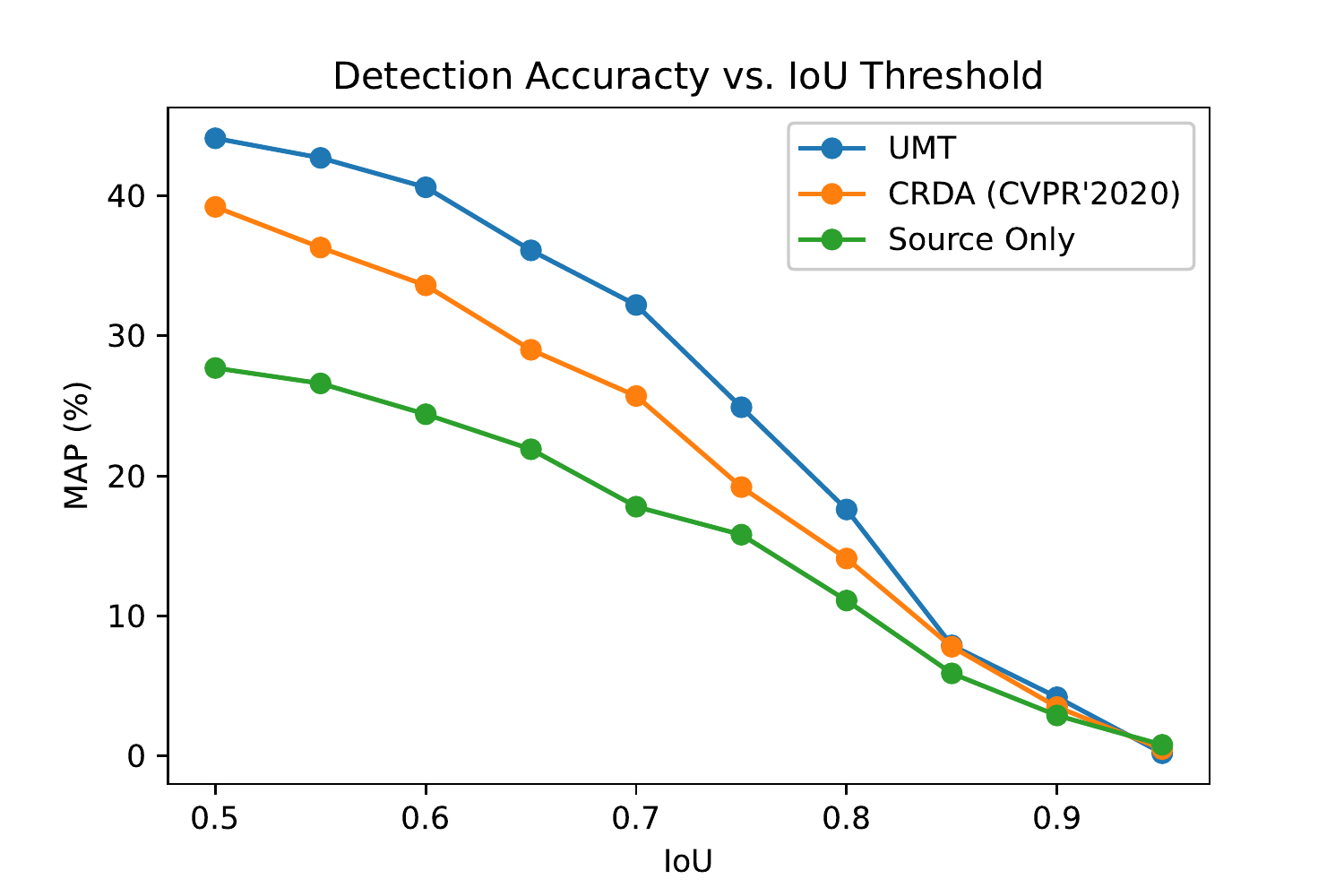}
  \caption{The performance with different of IoU thresholds on the adaptation task \textbf{PASCAL VOC$\rightarrow$Clipart1k}.}
  \label{fig:iou_map}
\end{figure}

\textbf{Influence of different IoU Thresholds:} In our experiments, following other state-of-the-art literatures~\cite{xu2020exploring,saito2019strong}, we set the IoU threshold as 0.5 to evaluate the model performance on the target domain. Fig.~\ref{fig:iou_map} illustrates the performance of different methods(\textit{i.e}, Source Only, CRDA~\cite{xu2020exploring} and our UMT) with the different IoU thresholds. We can observe that the MAP drops dramatically with the increasing of the IoU threshold due to more and more rigorous requirements of localization quality. We would like to highlight that the proposed UMT model consistently outperforms the counterpart~(\textit{i.e.}, CRDA and Source only) along with the IoU thresholds. This suggests that our UMT can detect objects with more robust and accurate bounding boxes regression.

\section{Qualitative Results}
\begin{figure*}[ht]
       \centering
         \resizebox{0.82\linewidth}{!}{%
          \includegraphics[width=1\linewidth]{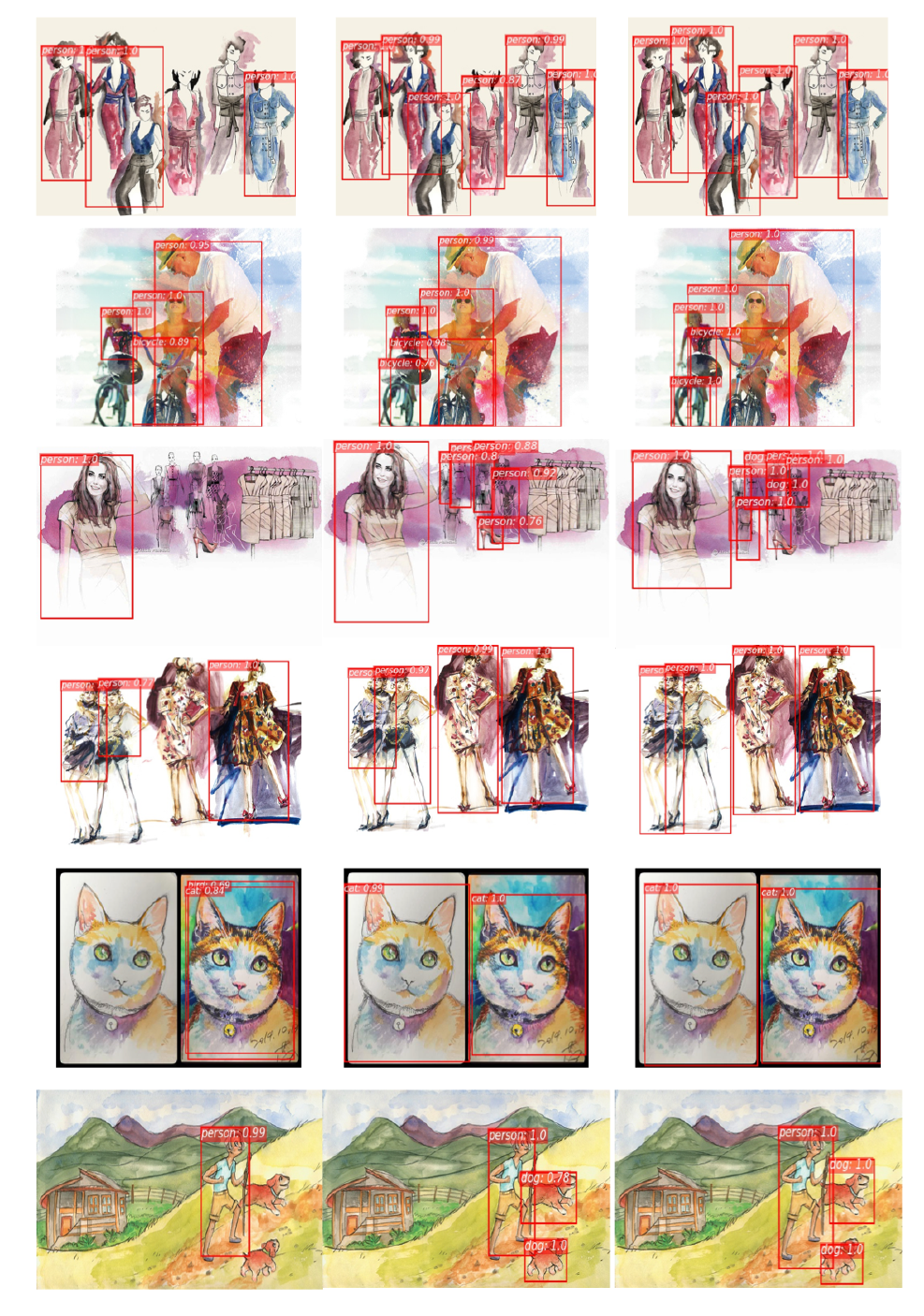}
         }
       \caption{Detection results visualization on Clipart1k for \textbf{PASCAL VOC$\rightarrow$Clipart1k} adaptation scene. From left to right, the results are Source Only, our UMT model, and ground truth. We show detections with category scores higher than $0.6$.} 
       \label{fig:vis_clipart}
     \end{figure*}

\begin{figure*}[ht]
         \centering
         \resizebox{0.70\linewidth}{!}{%
         \includegraphics[width=1\linewidth]{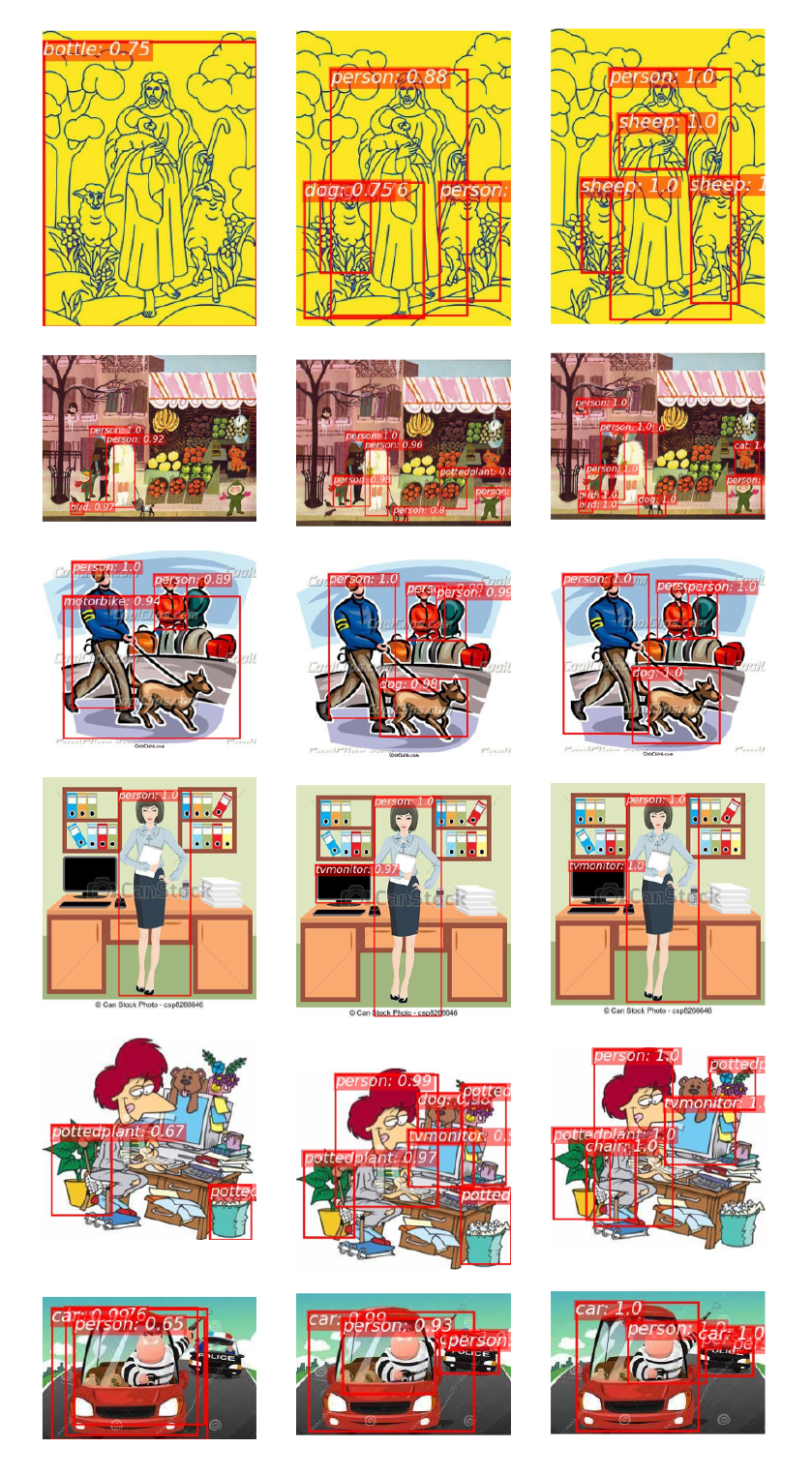}
         }
         \caption{Detection results visualization on Watercolor2k for \textbf{PASCAL VOC$\rightarrow$Watercolor2k} adaptation scene. From left to right, the results are Source Only, our UMT model, and ground truth. We show detections with category scores higher than $0.6$.} 
         \label{fig:vis_watercolor}
     \end{figure*}

\begin{figure*}[ht]
         \centering
         \includegraphics[width=1\linewidth]{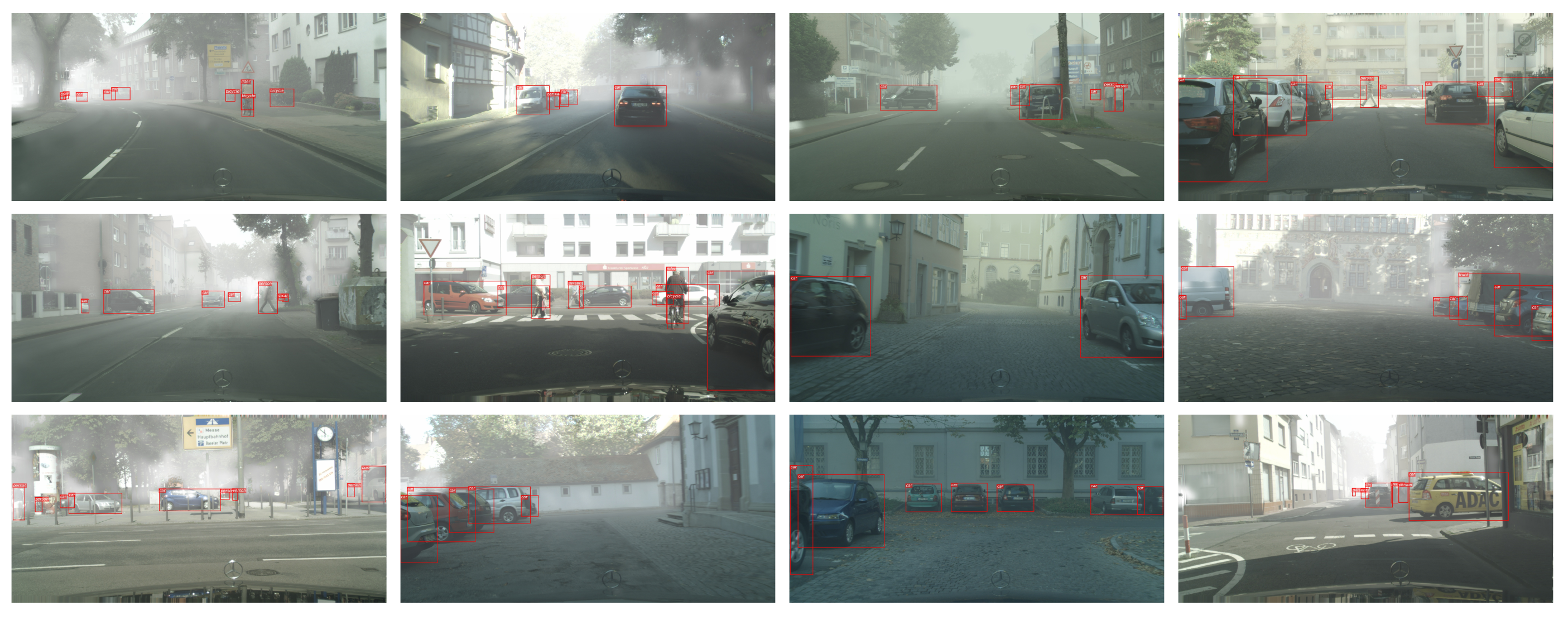}
         \caption{Detection results visualization on Foggy Cityscapes for \textbf{Cityscapes$\rightarrow$Foggy Cityscapes} adaptation scene. We can observed that the our $UMT$ model could accurately detect the objects in the image, e.g. person, aeroplane and motorbike. We show detections with the category scores higher than $0.6$.} 
         \label{fig:vis_foggy_cityscape}
     \end{figure*}

\begin{figure*}[ht]
         \centering
         \includegraphics[width=1\linewidth]{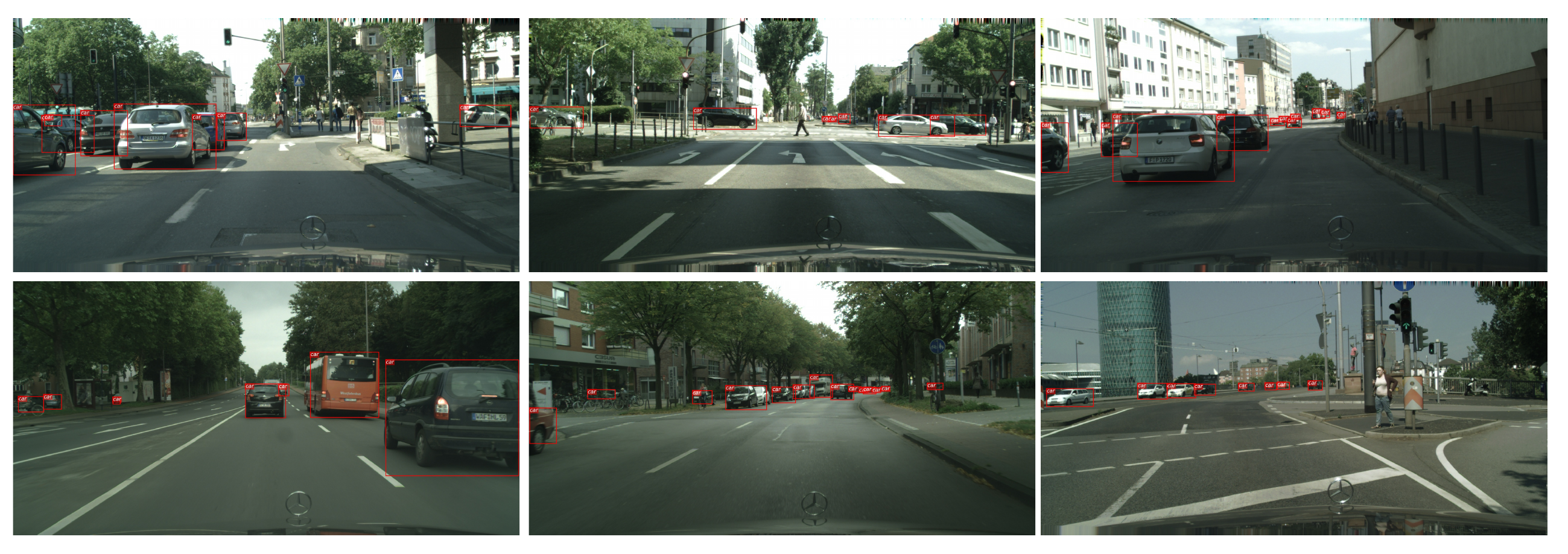}
         \caption{Detection results visualization on Cityscapes for \textbf{SIM10K$\rightarrow$Cityscapes} adaptation scene. We can observed that the our $UMT$ model could accurately detect the objects in the image (\textit{i.e.} car). We show detections with the category scores higher than $0.6$.} 
         \label{fig:vis_cityscape_car}
     \end{figure*}

    We also provide qualitative results for visual inspection. Sampled results on Clipart1k, Watercolor2k, Foggy Cityscapes and Cityscapes are illustrated in Fig.~\ref{fig:vis_clipart}, Fig.~\ref{fig:vis_watercolor}, Fig.~\ref{fig:vis_foggy_cityscape} and Fig.~\ref{fig:vis_cityscape_car} respectively. We can see that our UMT model could accurately detect the objects in the images.

\end{appendices}

\end{document}